\documentclass{article} % For LaTeX2e
\usepackage{conference,times}

\usepackage{amsmath,amsfonts,bm}

\def\eqref#1{equation~\ref{#1}}
\def\1{\bm{1}}

\DeclareMathAlphabet{\mathsfit}{\encodingdefault}{\sfdefault}{m}{sl}
\SetMathAlphabet{\mathsfit}{bold}{\encodingdefault}{\sfdefault}{bx}{n}

\usepackage{hyperref}
\usepackage{url}

\usepackage{booktabs}       % professional-quality tables
\usepackage{amsfonts}       % blackboard math symbols
\usepackage{nicefrac}       % compact symbols for 1/2, etc.
\usepackage{microtype}      % microtypography
\usepackage{xcolor}  
\usepackage{multirow} 
\usepackage{graphicx} 
\usepackage[table,dvipsnames]{xcolor}  % colors, including table row colors
\usepackage{wrapfig}
\usepackage{tikz}
\usetikzlibrary{arrows.meta,positioning,fit,backgrounds}

\usepackage{amsmath}

\newcommand{\metricpm}[1]{\textsubscript{\textcolor{gray}{\tiny $\pm$ #1}}}
\definecolor{lightgreen}{RGB}{236,250,236}
\definecolor{darkgreen}{RGB}{0,100,0}
\newcommand{\scoregain}[1]{\textcolor{darkgreen}{\textbf{#1}}}

\usepackage[scaled=0.9]{GoMono}
\newcommand{\ppol}{{\fontfamily{GoMono-TLF}\selectfont PPol}}

\usepackage{listings}

\usepackage{subcaption}

\definecolor{lightgreen}{RGB}{242,250,242}
\definecolor{listingbg}{RGB}{248,248,248}
\definecolor{listingframe}{RGB}{210,210,210}

\definecolor{basebg}{HTML}{fff0f0} % Very soft red tint for Base
\definecolor{ppolbg}{HTML}{E8F5E9} % Very soft green tint for PPol

\lstdefinestyle{appendixbox}{
  basicstyle=\ttfamily\scriptsize,
  breaklines=true,
  breakatwhitespace=false,
  columns=fullflexible,
  keepspaces=true,
  showstringspaces=false,
  frame=single,
  framerule=0.35pt,
  rulecolor=\color{listingframe},
  backgroundcolor=\color{listingbg},
  xleftmargin=0.5em,
  xrightmargin=0.5em,
  aboveskip=0.75em,
  belowskip=0.75em,
  literate={—}{{---}}1 {–}{{--}}1 {’}{{'}}1 {“}{{``}}1 {”}{{''}}1 {τ}{{$\tau$}}1 {²}{{$^2$}}1 {•}{{$\bullet$}}1
           {\{}{{\{}}1
           {\}}{{\}}}1
           {_}{{\_}}1
}

\usepackage{framed}
\usepackage{enumitem}
\usepackage{float}
\usepackage{fontawesome5}

\title{Beyond Cooperative Simulators: Generating Realistic User Personas for Robust Evaluation of LLM Agents}

\author{
\parbox{\linewidth}{
\vspace{2em}
\centering
Harshita Chopra$^{\dagger \clubsuit}$ \hspace{1em}
Kshitish Ghate$^{\dagger \clubsuit}$ \\[.2em]
Aylin Caliskan$^{\clubsuit}$ \hspace{1em}
Tadayoshi Kohno$^{\diamondsuit}$ \hspace{1em}
Chirag Shah$^{\clubsuit}$ \hspace{1em}
Natasha Jaques$^{\clubsuit}$ \\
\vspace{.4em}
{\normalfont $^\clubsuit$University of Washington, Seattle \quad $^\diamondsuit$Georgetown University, Washington DC} \\
\vspace{1.5em}
{\hspace{-1em}Code: \href{https://github.com/harshita-chopra/persona-policies}{\faGithub\ \textcolor{Blue}{Persona-Policies}}}
}
}

\iclrfinalcopy % Uncomment for camera-ready version, but NOT for submission.
\begin{document}

\vspace{-.2em}
\maketitle

\renewcommand{\thefootnote}{\fnsymbol{footnote}}
\footnotetext[2]{Primary contributors. Email: \{hchopra3, kghate\}@cs.washington.edu}

\begin{abstract}
Large Language Model (LLM) agents are increasingly deployed in settings where they interact with diverse users, including those who are unclear, impatient, or reluctant to share information. However, collecting real interaction data at scale remains expensive. The field has turned to LLM-based \emph{user simulators} as stand-ins, but these simulators inherit the behavior of their underlying models: cooperative and homogeneous. As a result, agents that appear strong in simulation often fail in real human interactions. To narrow this gap, we introduce \textbf{Persona Policies} (\ppol), a plug-and-play control layer that induces realistic behavioral variation in user simulators while preserving original task goals. Rather than hand-crafting personas, we employ an evolutionary coding agent to discover \emph{persona generation programs} optimized for human-likeness and behavioral coverage over real user conversations. The evolved program generates diverse, human-like personas for any task in the domain. Across 4 benchmarks---including $\tau^2$-bench Retail and Airline, ColBench, and WildChat---evolved \ppol{} yield 28--72\% absolute gains in fitness score over the baseline simulator. In blinded evaluations, annotators judged \ppol{} users as `\emph{human}' 80.4\% of the time, nearly 2$\times$ more than the baseline simulators. Training agents with \ppol{} also improves real-world performance: our user study with live human--agent interactions showed that fine-tuning with our method boosted task success by $+23\%$ over default baselines. \ppol{} thus offers a novel approach to strengthen simulator-based evaluation and training without changing underlying tasks.
\looseness=-1
\end{abstract}

\section{Introduction}
\label{sec:intro}

The promise of LLM-based assistants lies in their ability to act as autonomous agents, handling complex multi-turn tasks. However, to be truly useful, an agent must do more than navigate a software interface; it must navigate the unpredictable nature of human communication. Real users rarely provide information perfectly. They have distinct preferences, varying patience, and often express their needs through ambiguous, fragmented, or sometimes adversarial language \citep{keyvan2022approach}. To be trustworthy, agents must adequately infer intent, adapt to individual style, and ensure no irrevocable actions are taken without clear alignment.

Building agents with this level of social and procedural robustness requires evaluation environments that replicate the friction of real-world service encounters. The field has therefore turned to simulation-based benchmarks, which pair agents with ``user simulators'': LLMs prompted to act as users or customers with fixed goals across domains like retail, banking, and travel booking~\citep{barres2025tau2bench}. 
However, recent studies show that these default user simulators are overly cooperative, perfectly consistent, and highly forthcoming with information \citep{dou2025simulatorarena}. This creates a \emph{behavioral gap}: agents can appear strong against cooperative simulators but fail when confronted with the frustration, skepticism, or brevity of an actual human user \citep{zhou2026sim2real}.

Steering simulators with traits like impatience or skepticism~\citep{he2025impatient, shim2026noncollaborative} create a valuable stress test that lowers agent success rates. However, while ``richer synthetic users'' impact performance, they are typically hand-specified or sampled from fixed, manual designs rather than being grounded in the actual, complex distribution of real human behaviors.

\begin{figure*}[t]
  \centering
  \vspace{-1em}
  \includegraphics[width=1.0\textwidth]{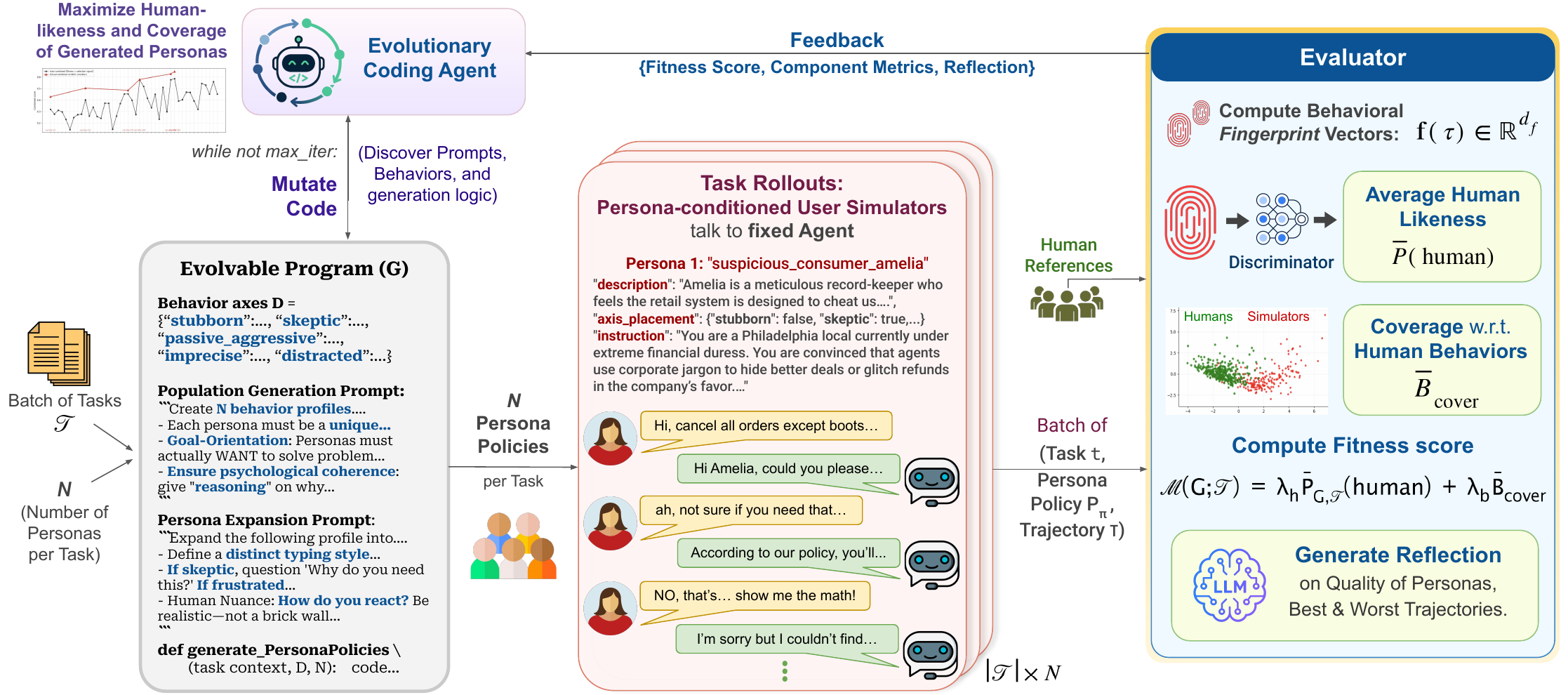}
  \caption{Overview of the \textbf{Persona Policies (\ppol)} program-evolution loop. Given a task, program $G$ generates $N$ persona policies that vary in how simulated users communicate. Candidate programs are scored for human-likeness and behavioral coverage via task rollouts. Feedback on the quality of trajectories is used to guide refinement of the axes, sampling rules, and prompts. Once optimized, program $G$ can produce diverse \textit{persona policies} for new tasks in that domain. 
  }
  \vspace{-.2em}
  \label{fig:evo}
\end{figure*}

To address this, we introduce \textbf{Persona Policies} (\ppol), a plug-and-play control layer that diversifies user simulation by generating multiple human-like personas while keeping benchmark tasks fixed. Each `\emph{policy}' is a short set of \emph{additional} instructions added to the simulator prompt that controls \emph{how} the user communicates, for example, by varying tone, pacing, selective disclosure, etc. This lets the same task be tested under many realistic interaction patterns, so benchmark performance better reflects robustness to real users rather than success on one cooperative simulator style.

Learning these policies is framed as \emph{evolutionary program search}. Rather than handcrafting prompts, we implement a Python persona generator that creates a population of behavioral profiles and expands them into task-preserving roleplay instructions. Using OpenEvolve~\citep{openevolve}, a frontier LLM iteratively mutates the generator's code. We evaluate candidate generators through agent--user rollouts scored on two objectives: \textbf{human-likeness}, measured as the probability of being classified as human by a trained discriminator, and \textbf{behavioral coverage}, which measures how well the $N$ personas span the human distribution. Both metrics rely on \emph{behavioral fingerprints}, 19 lexical and interaction-level features spanning communication style, information disclosure, clarification behavior, and error reaction~\citep{zhou2026sim2real}. A MAP-Elites archive~\citep{mouret2015illuminating} over these objectives keeps the population diverse rather than converging to a single behavioral type.

Beyond evaluation, evolved personas offer a powerful training signal. Agents fine-tuned on these interactions improve downstream robustness, analogous to domain randomization for language agents \citep{tobin2017domain}. In this work, we make three contributions:
\begin{enumerate}[topsep=-2pt,itemsep=0pt]
  \item We design \ppol{} as a plug-and-play control layer that injects realistic behavioral diversity into standard user simulators without changing the underlying tasks.

  \item We formulate persona generation as an automated framework using evolutionary search to discover diverse, human-like behaviors without manual intervention.

  \item We validate \ppol{} end-to-end via user studies: showing evolved personas appear largely human-like and training agents with \ppol{} yields higher success in real human interactions.
\end{enumerate}

\vspace{-.05em}
\section{Related Work}
\label{sec:relatedwork}
\vspace{-.3em}

\textbf{Multi-turn agentic tasks.}
LLMs are increasingly evaluated in dynamic, multi-turn tool environments rather than on static prompts. Benchmarks like MINT \citep{wang2024mint} and ToolSandbox \citep{lu2025toolsandbox} evaluate iterative tool use and stateful rollouts, while $\tau^2$-bench \citep{barres2025tau2bench} formalizes policy-constrained interactions in shared environments. However, even capable models degrade significantly when information is revealed incrementally across turns, underscoring the fundamental difficulty of multi-turn execution \citep{laban2026llms}.

\textbf{User simulation.}
Prior work aims to make simulated interactions more coherent, goal-consistent and natural. LLM-based simulators improve task-oriented dialogue via fine-tuning, verifiers, or specialized questioning patterns \citep{sekulic2024reliable,luo2024duetsim,kong2024platolm}. A parallel line of persona-based work uses profiles to diversify generated user responses \citep{zhang-etal-2018-personalizing,ge2024scaling,wang-etal-2025-know}, though typically treating personas as static prompt artifacts. Recent work \citep{paglieri2026persona} introduced an approach to automatically evolve persona generators to maximize socio-behavioral diversity in responses. Their focus, however, remains on populating open-ended, game-like engine environments (e.g., Concordia), instead of structured, task-grounded interaction. In contrast,\ppol{} optimizes persona evolution directly from multi-turn agent rollouts: given a fixed task, it generates a population of users who all pursue the same goal but communicate in different ways. Our objective ensures that the resulting conversations resemble real human interactions and the generated population covers diverse regions of the human behavior distribution.

\textbf{Robustness under diverse users.}
% A growing line of work frames robustness in interactive systems as exposure to diverse user behaviors. Simulators that inject traits like impatience, ambiguity, or non-cooperation can significantly degrade success rates \citep{he2025impatient,shim2026noncollaborative}. While these studies establish that richer synthetic users alter measured outcomes, such traits are typically hand-specified or sampled from fixed templates \citep{dou2025simulatorarena,seshadri2026lost,zhou2026sim2real}. This limits their ability to systematically probe how agents respond to the full, complex distribution of real human interaction. \ppol{} addresses this by automating the discovery and scaling of these diverse behaviors.
Evaluation scores on interactive benchmarks reflect \emph{both} the agent and the simulated user who conversed. Recent meta-evaluations show that switching simulators can reorder model rankings and diverge from human judgments \citep{dou2025simulatorarena,seshadri2026lost,zhou2026sim2real}. Separately, stress-test studies show that small, controlled changes to user behavior, such as prompting more impatient or less cooperative play, can move headline success rates by large margins \citep{he2025impatient,shim2026noncollaborative}. Together, these findings argue for making the simulated user an explicit, tunable part of the benchmark, rather than a hidden default baked into the evaluation. To this end, the \ppol{} framework discovers user-side interaction behaviors and keeps selection grounded in signals computed from real human traces.

%%%%%%%%%%%%%%%%%%%%%%%%%%%%%%%%%%%%%%%%%%%%%%
\section{Method}
\label{sec:method}
\vspace{-.5em}

\textbf{Persona Policies (\ppol)} automate the creation of a domain-level program that generates multiple task-preserving personas for any given user scenario. We first state the setting and notation, then describe the generator, fitness score, evolutionary search, and implementation.

\vspace{-.3em}
\subsection{Problem Setting}
\label{sec:method-setting}
\vspace{-.3em}

We consider multi-turn, task-oriented dialogue where an LLM \emph{user simulator} interacts with an \emph{agent}. Each task $t$ provides a base context $c_t$ containing the user's scenario, goals, and private facts. To maintain standard benchmark evaluation, all underlying task elements: environment state, task specification, and success criteria, remain fixed. We control only the simulator's behavioral style by appending natural-language persona instructions to its system prompt. 

\textbf{Persona policy}~$P_\pi$ refers to appended natural-language text that controls the simulated user's communicative style, including tone, pacing, and disclosure, while preserving task objectives in $s_{\mathrm{base}}(t)$, the benchmark's default system prompt for task~$t$. The final user-side system prompt becomes $s_{\mathrm{base}}(t) \oplus P_\pi$, where $\oplus$ denotes string concatenation.

\begin{figure*}[t]
  \centering
  \vspace{-1em}
      \includegraphics[width=1.0\textwidth]{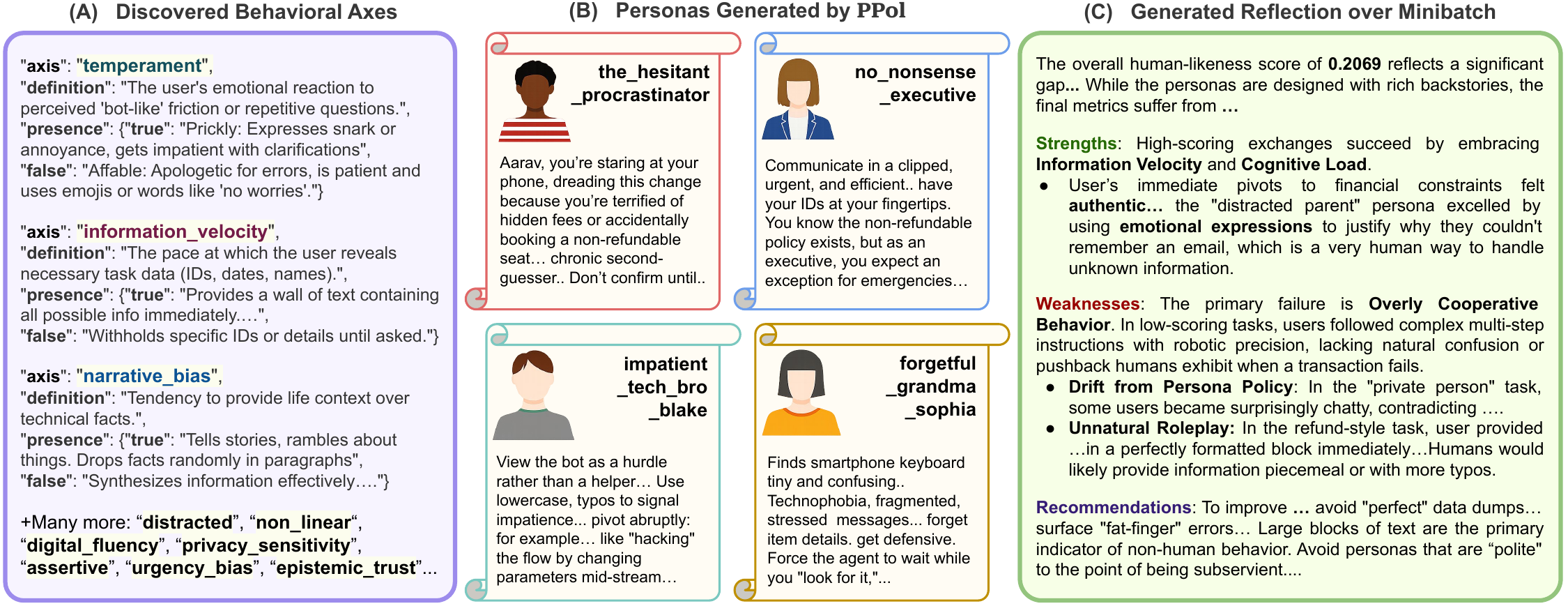}
      \caption{Samples of (A) Discovered Behavioral Axes (drives diversity in population), (B) Personas generated by \ppol{}, (C) Reflection over a minibatch (provided as feedback to mutate program).
      }
  \label{fig:samples}
\end{figure*}

Behavioral variation is steered by a list of \textbf{behavioral axes}, $D$, where each axis includes a short definition with dual-state \emph{playbooks} for how the behavior appears when active or inactive (e.g., Figure~\ref{fig:samples}A). We seed $D$ with four behaviors: terseness, skepticism, frustration, and ambiguity. \looseness=-1
The list~$D$  resides inside the evolvable generator program, allowing mutation to modify the axes alongside prompt instructions. Our evolutionary loop has four main steps: (1)~generate persona policies per task, (2)~run agent--user rollouts, (3)~score trajectories against human behavior, and (4)~mutate the generator code. 

\subsection{Persona generation}
\label{sec:method-generator}
\vspace{-.3em}

The evolvable artifact is a Python program that implements the persona generator function~$G(c_t,D,N)$, which outputs a set of $N$ persona policies given the task context~$c_t$:
\vspace{-.5em}
\begin{equation}
\label{eq:generator_output}
\Pi_t
=
\left\{
P_{\pi,t}^{(i)}
\right\}_{i=1}^{N},
\qquad
\Pi_t = G(c_t,D,N)
\end{equation}

\vspace{-.5em}
Each persona policy~$P_{\pi,t}^{(i)}$ has a \textbf{persona record}
\(
r_t^{(i)}=\left(\mathbf{a}_t^{(i)},P_{\pi,t}^{(i)}\right).
\)
Here $\mathbf{a}_t^{(i)} \in \{0,1\}^{|D|}$ is the binary mapping that determines which behavioral axes are active (true) for persona~$i$ on task~$t$.

\vspace{-.5em}
\paragraph{Population generation.}
The \textbf{first phase} is \emph{population generation}: a frontier language model, conditioned on the scenario~$c_t$ and full specification of~$D$, jointly proposes~$N$ \emph{population members}. The structured response lists exactly~$N$ members, each with a short natural-language summary, brief rationale, and axis assignment in $\{0,1\}^{|D|}$. Because assignments are chosen jointly, this phase drives \textbf{diversity} in the discrete behavior space before any long-form user-facing policy is written.

\vspace{-.5em}
\paragraph{Persona expansion.}
The \textbf{second phase} \emph{expands} each population member, independently and in parallel, into one long natural-language persona policy~$P_{\pi,t}^{(i)}$, the string appended to $s_{\mathrm{base}}(t)$ for that rollout. Each call is conditioned on~$c_t$, ~$\mathbf{a}_t^{(i)}$, and the active-axis playbooks.

\subsection{Optimization Metrics}
\label{sec:method-fitness}
We score a candidate generator~$G$ by running it and evaluating the resulting dialogue trajectories. For each task~$t$ in a minibatch~$\mathcal{T}$, we generate $\Pi_t = G(c_t,D,N)$ and run a full rollout for each policy~$P_{\pi,t}^{(i)} \in \Pi_t$, with the persona-conditioned user simulator talking to the fixed agent. This gives one completed trajectory per $(t,i)$ pair. We convert each trajectory into a behavioral fingerprint, explained below, and score the minibatch along two axes. First, \textbf{human-likeness} is the average probability that simulated trajectories resemble real user trajectories in the defined feature space. Second, \textbf{behavioral coverage}, a reward encouraging the generated personas to spread over the human-like behavior distribution rather than collapsing to one stereotyped style.

\paragraph{Human reference and behavioral fingerprints.} We compare LLM simulators to real users on multi-turn tasks along four axes: communication style (D1), information disclosure (D2), clarification behavior (D3), and error reaction (D4), following \citep{zhou2026sim2real}. We create a vector with $d_f = 19$ scalar features computed only from \emph{user} turns in a completed trajectory~$\tau$. Each feature is a rate, count, or normalized statistic derived from regular-expression patterns over user messages (politeness, uncertainty, pushback, etc.) and simple turn statistics (length, variability, repetition) utilizing the LIWC2015~\citep{pennebaker2015development} and NRC~\citep{mohammad2013crowdsourcing} lexicons. Examples include \emph{words per turn} and \emph{short-utterance rate} (D1), \emph{front-loading ratio} of identifying information (D2), \emph{clarification-question rate} and \emph{pushback rate} (D3), and \emph{emotional-expression rate} (D4). A full list of the 19 features is provided in Appendix~\ref{app:fingerprint-features}. Stacking these scalars yields the episode's \emph{behavioral fingerprint}, $\mathbf{f}(\tau) \in \mathbb{R}^{d_f}$. 
We use a fixed calibration corpus of human conversations, $\mathcal{H}$, to compute the elementwise mean fingerprint $\boldsymbol{\mu}_{\mathcal{H}} \in \mathbb{R}^{d_f}$ and build the discriminator training set.

\paragraph{Human-likeness via a learned discriminator.}
We train a lightweight binary classifier (a Random Forest; see Appendix~\ref{app:discriminator-details} for performance analyses) to separate fingerprint vectors  
% from~$\mathcal{H}$ (label \emph{human}) from fingerprints collected under the default cooperative user simulator on the same benchmark configuration (label \emph{simulator}). 
from real human dialogues in~$\mathcal{H}$ (label \emph{human}) and fingerprint vectors collected under the LLM-based default user simulator on the same benchmark configuration (label \emph{simulator}).
Inputs are standardized before fitting. At evaluation time, for an episode~$e$ with fingerprint $\mathbf{f}_e$, the model outputs an estimated probability $p_{\mathrm{RF}}(\mathrm{human}\mid \mathbf{f}_e)$ that the trajectory looks human-like in this feature space. For a generator~$G$ evaluated on task minibatch~$\mathcal{T}$, let $\mathcal{B}(G;\mathcal{T})$ be the resulting set of rollout episodes. We aggregate over this set, and define our scalar signal for discriminator-based human-likeness as: \looseness=-1
\begin{equation}
\label{eq:human_likeness}
\overline{P}_{G,\mathcal{T}}(\mathrm{human}) \;=\; \frac{1}{|\mathcal{B}(G;\mathcal{T})|} \sum_{e \in \mathcal{B}(G;\mathcal{T})} p_{\mathrm{RF}}(\mathrm{human}\mid \mathbf{f}_e)
\end{equation}

\textbf{Behavioral coverage.}
Let $\mathcal{F}_t(G)=\{\mathbf{f}^{(1)},\ldots,\mathbf{f}^{(N)}\}$  $\forall$ $\mathbf{f}^{(i)} \in \mathbb{R}^{d_f}$ be the set of fingerprint vectors of the~$N$ personas derived from their independent rollout performing task $t$. Let $\mathcal{H}_{\mathrm{ref}}$ denote the set of fingerprint vectors from the real human reference data (dialogue traces) in the target domain. Considering $\mathcal{F}_t(G)$ and $\mathcal{H}_{\mathrm{train}}$ as two finite point clouds in $\mathbb{R}^{d_f}$, their two-sided Chamfer error is
\[
\mathrm{err}(\mathcal{F}_t(G),\mathcal{H}_{\mathrm{ref}}) \;=\; \frac{1}{|\mathcal{H}_{\mathrm{ref}}|} \sum_{h \in \mathcal{H}_{\mathrm{ref}}} \min_{\mathbf{f} \in \mathcal{F}_t(G)} \|h - \mathbf{f}\|_2 \;+\; \frac{1}{|\mathcal{F}_t(G)|} \sum_{\mathbf{f} \in \mathcal{F}_t(G)} \min_{h \in \mathcal{H}_{\mathrm{ref}}} \|\mathbf{f} - h\|_2
\]
where $\|\cdot\|_2$ denotes the standard Euclidean ($L_2$) norm. To map this metric on a $[0,1]$ scale comparable to human-likeness, let $d_{\mathrm{ref}}$ be the mean pairwise distance among points in~$\mathcal{H}_{\mathrm{ref}}$, representing how spread out real users are in fingerprint space. For $N \geq 1$, the per-task \emph{coverage} score is
\begin{equation}
\label{eq:b_cover_task}
B_{\mathrm{cover}}(\mathcal{F}_t(G),\mathcal{H}_{\mathrm{ref}}) \;=\; \max\!\left\{0,\, 1 - \min\!\left(1,\, \frac{\mathrm{err}(\mathcal{F}_t(G),\mathcal{H}_{\mathrm{ref}})}{2\, d_{\mathrm{ref}}}\right)\right\},
\end{equation}
Intuitively, Chamfer score measures how well two point clouds sit on top of each other: for every human point, how far is the nearest persona (coverage), and for every persona point, how far is the nearest human (realism)? 
Optimizing this ensures persona fingerprints stay near the support of the human reference distribution. Averaging over tasks in the minibatch gives
\begin{equation}
\label{eq:b_cover_mean}
\overline{B}_{\mathrm{cover}}(G;\mathcal{T}) \;=\; \frac{1}{|\mathcal{T}|}\sum_{t\in\mathcal{T}} B_{\mathrm{cover}}(\mathcal{F}_t(G),\mathcal{H}_{\mathrm{ref}})
\end{equation}

\textbf{Combined score.}
Relying solely on a discriminator trained against default simulators risks adversarial drift, where the generator could produce unnatural behaviors that exploit the classifier's blind spots. The behavioral coverage term ($\overline{B}_{\mathrm{cover}}$) acts as a geometric regularizer against this: if the evolved personas drift off the human support in the fingerprint space, their Chamfer distance to the human reference cloud increases, heavily penalizing the combined fitness.
The quantity that drives selection during evolution is the batch fitness
\begin{equation}
\label{eq:combined_score}
\mathcal{M}(G;\mathcal{T}) \;=\; \lambda_h \, \overline{P}_{G,\mathcal{T}}(\mathrm{human}) \;+\; \lambda_b \, \overline{B}_{\mathrm{cover}}(G;\mathcal{T}), \qquad \lambda_h + \lambda_b = 1.
\end{equation}

\subsection{Evolutionary search}
\label{sec:method-openevolve}
\vspace{-.1em}
Search is implemented with OpenEvolve~\citep{openevolve}: the Python program implementing~$G$ is the genotype, and a mutation language model proposes edits to prompts, the axis list~$D$, and control flow. Each outer iteration samples a training minibatch~$\mathcal{T}$ using sliding windows over the task pool. We use a curriculum that increases the persona count~$N$ across epochs (e.g., $5 \rightarrow 8 \rightarrow 10$), so early iterations emphasize human-likeness under a smaller persona set while later iterations increase pressure on coverage. The coverage weight $\lambda_b$ increases with the ratio of the current training~$N$ to the terminal~$N$, and $\lambda_h = 1 - \lambda_b$. For the current program, we reload the module, run generation and expansion for every task in~$\mathcal{T}$, execute benchmark rollouts for each $(\mathrm{task},\,\mathrm{persona})$ pair, compute fingerprints, and evaluate Equations \eqref{eq:human_likeness}--\eqref{eq:combined_score}. The resulting fitness~$\mathcal{M}(G;\mathcal{T})$ and the two-dimensional behavioral coordinates $\bigl(\overline{P}_{G,\mathcal{T}}(\mathrm{human}),\, \overline{B}_{\mathrm{cover}}(G;\mathcal{T})\bigr)$ are returned to OpenEvolve; MAP-Elites~\citep{mouret2015illuminating} partitions programs into bins over these coordinates and retains the highest-$\mathcal{M}$ program per bin, maintaining a diverse archive of behavioral strategies. We also evaluate a held-out validation slice of the training pool for monitoring.

\paragraph{Reflection-guided mutation.}
Beyond $\mathcal{M}(G;\mathcal{T})$, we construct a natural-language \emph{reflection} artifact to guide the mutator. A separate reflection model analyzes fitness summaries ($\mathcal{M}$, mean human-likeness, mean coverage) alongside trajectory excerpts from the best and worst rollouts ranked by~$p_{\mathrm{RF}}(\mathrm{human}\mid\mathbf{f})$. Each example includes the persona policy, the 19-D fingerprint vector, and the dialogue. The model returns a concise critique identifying which dialogue patterns plausibly drive high or low human-likeness (e.g., see Figure~\ref{fig:samples}(C)). This reflection is passed to the OpenEvolve mutation step with its system templates, keeping the optimization anchored to the score in Eq.\eqref{eq:combined_score}.

\begin{figure}[t]
  \centering
  \vspace{-1em}
      \includegraphics[width=.98\textwidth]{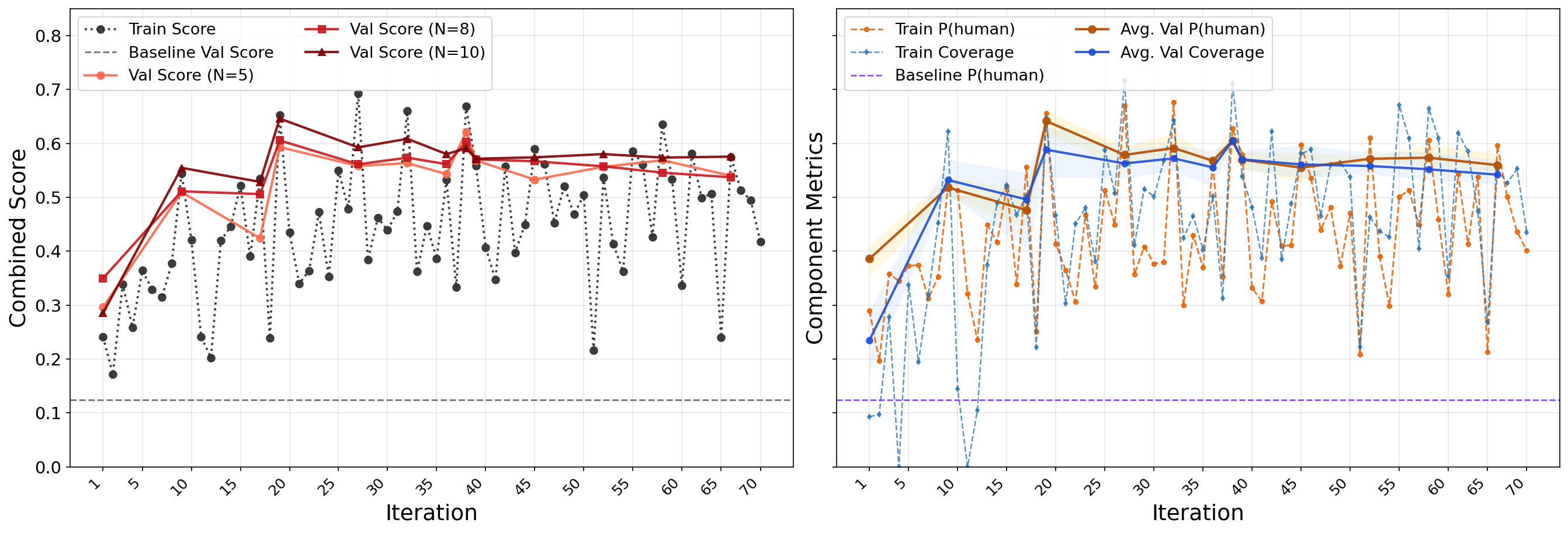}
  \vspace{-.5em}
  \caption{Evolution metrics on $\tau^2$-bench Retail for $N \in \{5, 8, 10\}$ personas. Combined score (left) and averaged component metrics (right) improve concurrently, outperforming the baseline.
  % Domain: Retail. User Simulator: Qwen3-Next-80B-A3B-Instruct.
  }
  \label{fig:evolution-curves}
\end{figure}

%%%%%%%%%%%%%%%%%%%%%%%%%%%%%%%%%%%%%%%%%%%%%%
\section{Experiments}
\label{sec:experiments}
\vspace{-.2em}

First, we compare the quality of responses generated by \ppol{} and other baselines with real users. We then validate the behavioral fingerprint-based scores with a blinded human study. We also designed a live multi-turn interaction platform to test the performance of agents trained on \ppol{} rollouts.

\vspace{-.5em}
\paragraph{Datasets.}
We evaluate \ppol{} on 4 popular benchmarks. \textbf{Retail} and \textbf{Airline} domains of $\tau^2$-bench~\citep{barres2025tau2bench} represent customer-support tasks. We use the 495 corresponding real user trajectories from \citet{zhou2026sim2real} as reference data for human behavioral fingerprints ($\mathcal{H}_{\mathrm{ref}}$).
\textbf{ColBench} \citep{zhou2025sweet} is a dataset of multi-turn collaborative coding scenarios; we use the \emph{backend programming} tasks. For its human reference data, we use SWE-Chat \citep{baumann2026swechat} dialogues, filtered for English and code-authoring intents, yielding 1,449 real developer–agent dialogues across 63 repos. \textbf{RealUserSim} \citep{zhu2026realusersim} dataset comprises 600 real WildChat \citep{zhao2024wildchat} user–AI conversations spanning 72 domains across 460 users, each paired with an extracted problem description to ground the user simulator. Dataset splits are reported in Appendix~\ref{app:datasets}.

\vspace{-.5em}
\paragraph{Models.}
We evaluate three user-simulator backends: DeepSeek-V3.1, Qwen3-Next-80B-A3B-Instruct, and GPT-5.4-Mini~\citep{deepseekai2024deepseekv3technicalreport, qwen3technicalreport, gpt54mini2026}. The assistant, environment interface, and natural-language checks use Gemma~4 (31B, instruction-tuned). Evolutionary program search uses Gemini~3 Flash for generator calls, reflection, and OpenEvolve code-edit proposals. All models are invoked through a unified LiteLLM interface, and rollouts are parallelized during training. Initial generator and reflection prompt are shown in Appendices~\ref{app:initial-generator} and~\ref{app:reflection-prompt}.

\vspace{-.5em}
\paragraph{Baselines.}
We compare 4 simulator variants.
\emph{Base-simulator} uses the default benchmark-provided user prompt with no added personas.
\emph{Direct-Prompt (DP) Personas} uses a single LLM call per task to generate~$N{=}10$ personas altogether.
\emph{\ppol{}-Initial} uses the unevolved seed generator.
\emph{\ppol{}-Evolved} uses the best evolved checkpoint, selected by mean validation score across $N{\in}\{5,8,10\}$ over a maximum budget of 70 iterations.
All \ppol{} variants generate $N{=}10$ personas per task at test time. 
\textbf{Humans'} behavior is included as a reference row. Coverage here is typically ${<}1$, as it measures how well a group of real human dialogues spreads across the full human reference distribution. $\tau^2$ data from ~\citet{zhou2026sim2real} provides three human dialogs per task, which we use directly as this group. ColBench and WildChat have only one human dialogue per task, so we cluster similar tasks and average the coverage of $N{=}10$-sized subsamples of human dialogues from each cluster (Appendix~\ref{app:human-coverage}). 
\looseness=-1

\vspace{-.5em}
\paragraph{Metrics.}
We report human-likeness (HL), behavioral coverage, and the combined score from Section~\ref{sec:method-fitness}. To complement HL, which measures how human-like a trajectory appears to the discriminator, we also report per-dimension S{\o}rensen--Dice alignment ($D_1$--$D_4$) and the aggregate $\mathrm{USI}_{\mathrm{D1\text{-}D4}}$, the mean of the four dimension-level Dice coefficients. Dice alignment (higher$=$closer to human) measures similarity to the human reference along specific behavioral dimensions by comparing the mean generated fingerprint to the human reference mean~\citep{zhou2026sim2real}.

% \paragraph{Interactive User Study.}
% The ultimate test of a user simulator is whether agents trained on it perform better with real humans. However, evaluating human transfer in benchmarks like $\tau^2$-bench is challenging because they rely on highly specified scenarios (e.g., requesting a shoe exchange, then changing mind to ask for a refund) that are hard to replicate with real humans without forcing them into rigid, artificial scripts. To address this, we designed a novel multi-turn user study protocol (Appendix~\ref{app:train-human-eval}) that dynamically feeds instructions, constraints, and facts to human participants via real-time guidance flashcards as the conversation unfolds. This forces users to adhere to benchmark goals while granting them complete freedom to phrase, hesitate, or react naturally. We leveraged this protocol to conduct a double-blind trial ($N = 24$ participants across Retail and Airline scenarios), where each user completed 10 interactive chats randomly split between agents trained on \ppol{} evolved personas versus the baseline simulator.

\vspace{-.5em}
\paragraph{Interactive User Study.}
The ultimate test of a user simulator is whether agents trained on it perform better with real humans. However, evaluating human transfer in benchmarks like $\tau^2$-bench is challenging because they rely on highly specified scenarios (e.g., requesting a shoe exchange, then changing mind to ask for a refund) that are hard to replicate with real humans without forcing them into rigid, artificial scripts. To address this, we designed a novel multi-turn user study protocol (Appendix~\ref{app:train-human-eval}) that dynamically feeds instructions, constraints, and facts to human participants as the conversation unfolds. To elicit authentic human behavior while still adhering to benchmark goals, and to prevent scripted, upfront information dumping, the interface incorporates three core mechanisms: \looseness=-1
\begin{itemize}[leftmargin=2em, itemsep=-.2em, topsep=-.2em]
    \item \textbf{Dynamic Direction Flashcards:} An online LLM tracks conversation progress in real time and updates suggestion cards on the right panel. These flashcards present context-aware directions, alternative choices, and behavioral cues that participants can optionally adopt at each turn.
    \item \textbf{Incremental Information Disclosure:} Rather than presenting all scenario facts upfront, details are revealed piece-by-piece as the conversation progresses to simulate natural memory recall and realistic information sharing.
    \item \textbf{Live Chat Summary:} A dynamic summary pane automatically updates as information is shared, keeping participants aware of what has already been disclosed by them in the current chat.
\end{itemize}
This enables authentic multi-turn interactions by balancing user expression with task constraints. In a double-blind evaluation across 30 Retail and Airline tasks, 36 participants completed 239 chats across two agent conditions: one trained with \ppol{} versus another with the base-simulator. \looseness=-1

%%%%%%%%%%%%%%%%%%%%%%%%%%%%%%%%%%%%%%%%%%%%%%
\vspace{-.2em}
\section{Results \& Discussion}

\begin{table*}[t]
  \centering
  \scriptsize
  \caption{\textbf{Results.}
  HL is the mean discriminator probability $\overline{P}(\mathrm{human})$, Coverage is the mean behavioral coverage $\overline{B}_{\mathrm{cover}}$, and Score is the fitness $\mathcal{M}$ from \eqref{eq:combined_score} reported on a $[0,1]$ scale.
  $D_1$--$D_4$ are S{\o}rensen--Dice alignment scores for communication style, information disclosure, clarification behavior, and error reaction, respectively. Values are reported as (mean$\pm{\text{std}}$). 
  % Human Coverage is \(<1\) because it compares a small human subset to the broader reference distribution.
  }
  \vspace{-.2em}
  \label{tab:persona_benchmark}
  \resizebox{\linewidth}{!}{%
  \begin{tabular}{l c c c c c c c c}
    \toprule
    & \multicolumn{3}{c}{Optimization Metrics $\uparrow$} & \multicolumn{5}{c}{Behavioral Alignment (\%) $\uparrow$} \\
    \cmidrule(lr){2-4} \cmidrule(lr){5-9}
    Method & HL & Coverage  & Score  & $D_1$  & $D_2$  & $D_3$  & $D_4$  & $\mathrm{USI}_{\mathrm{D1-D4}}$ \\
    
    \midrule
    \multicolumn{9}{l}{Dataset: \textbf{$\tau^2$Retail}, User Simulator: Qwen3-Next-80B-A3B-Instruct} \\
    \hline

    \rowcolor{lightgreen} Humans & 0.953 \metricpm{0.076} & 0.614 \metricpm{0.091} & 0.783 \metricpm{0.055} & 84.3 \metricpm{2.3} & 97.6 \metricpm{2.2} & 87.2 \metricpm{6.0} & 82.2 \metricpm{12.6} & 87.8 \metricpm{3.5} \\

    Base-simulator & 0.107 \metricpm{0.244} & 0.046 \metricpm{0.136} & 0.077 \metricpm{0.186} & 24.0 \metricpm{2.4} & 57.5 \metricpm{5.8} & 35.7 \metricpm{6.3} & 23.3 \metricpm{6.6} & 35.1 \metricpm{2.9} \\

    DP Personas & 0.291 \metricpm{0.355} & 0.100 \metricpm{0.219} & 0.196 \metricpm{0.212} & 29.5 \metricpm{1.8} & 56.5 \metricpm{2.8} & 37.8 \metricpm{3.5} & 25.2 \metricpm{3.1} & 37.2 \metricpm{2.0} \\

    \ppol{}: Initial & 0.356 \metricpm{0.348} & 0.017 \metricpm{0.085} & 0.186 \metricpm{0.083} & 31.4 \metricpm{2.1} & 58.7 \metricpm{2.7} & 38.7 \metricpm{2.3} & 30.3 \metricpm{2.4} & 39.8 \metricpm{1.4} \\

    \textbf{\ppol{}: Evolved} & \textbf{0.784 \metricpm{0.204}} & \textbf{0.602 \metricpm{0.161}} & \textbf{0.693 \metricpm{0.139}} & \textbf{69.6 \metricpm{5.0}} & \textbf{89.8 \metricpm{2.4}} & \textbf{70.4 \metricpm{4.5}} & \textbf{76.4 \metricpm{8.1}} & \textbf{76.5 \metricpm{2.6}} \\

    \midrule
    \multicolumn{9}{l}{Domain: \textbf{$\tau^2$Airline + Retail}, User Simulator: DeepSeek-V3.1} \\
    \hline

    \rowcolor{lightgreen} Humans & 0.958 \metricpm{0.073} & 0.623 \metricpm{0.089} & 0.790 \metricpm{0.051} & 94.9 \metricpm{4.7} & 97.8 \metricpm{1.8} & 88.6 \metricpm{5.2} & 92.2 \metricpm{8.3} & 93.4 \metricpm{3.0} \\

    Base-simulator & 0.123 \metricpm{0.177} & 0.146 \metricpm{0.176} & 0.135 \metricpm{0.159} & 32.0 \metricpm{2.6} & 88.7 \metricpm{2.8} & 53.4 \metricpm{5.8} & 54.4 \metricpm{10.3} & 57.1 \metricpm{3.2} \\

    DP Personas & 0.292 \metricpm{0.317} & 0.340 \metricpm{0.264} & 0.316 \metricpm{0.203} & 37.1 \metricpm{1.1} & 75.9 \metricpm{1.9} & 49.8 \metricpm{2.4} & 32.9 \metricpm{2.0} & 48.9 \metricpm{1.2} \\

    \ppol{}: Initial & 0.410 \metricpm{0.332} & 0.323 \metricpm{0.210} & 0.367 \metricpm{0.139} & 42.7 \metricpm{1.3} & 85.3 \metricpm{2.6} & 43.8 \metricpm{2.8} & 33.2 \metricpm{2.0} & 51.2 \metricpm{1.4} \\

    \textbf{\ppol{}: Evolved} & \textbf{0.570 \metricpm{0.313}} & \textbf{0.657 \metricpm{0.116}} & \textbf{0.614 \metricpm{0.133}} & \textbf{65.4 \metricpm{3.3}} & \textbf{93.7 \metricpm{2.2}} & \textbf{84.0 \metricpm{5.3}} & \textbf{54.8 \metricpm{6.0}} & \textbf{74.5 \metricpm{2.2}} \\

    \midrule
    \multicolumn{9}{l}{Dataset: \textbf{ColBench}, User Simulator: GPT-5.4-Mini} \\
    \hline

    \rowcolor{lightgreen} Humans & 0.929 \metricpm{0.169} & 0.694 \metricpm{0.022} & 0.811 \metricpm{0.120} & 79.3 \metricpm{1.8} & 92.9 \metricpm{1.9} & 66.4 \metricpm{5.2} & 65.2 \metricpm{5.0} & 75.9 \metricpm{2.0} \\
  
      Base-simulator & 0.307 \metricpm{0.268} & 0.574 \metricpm{0.081} & 0.440 \metricpm{0.089} & 48.9 \metricpm{5.5} & 56.6 \metricpm{2.3} & 41.6 \metricpm{5.6} & 20.2 \metricpm{6.3} & 41.8 \metricpm{2.7} \\
  
      DP Personas & 0.542 \metricpm{0.306} & 0.347 \metricpm{0.125} & 0.444 \metricpm{0.075} & 54.7 \metricpm{4.2} & 57.1 \metricpm{2.7} & 59.1 \metricpm{3.9} & 47.4 \metricpm{9.1} & 54.6 \metricpm{2.9} \\
  
      \ppol{}: Initial & 0.659 \metricpm{0.306} & 0.418 \metricpm{0.104} & 0.539 \metricpm{0.070} & 64.2 \metricpm{4.0} & \textbf{59.2 \metricpm{2.5}} & \textbf{63.6 \metricpm{3.3}} & 49.0 \metricpm{7.2} & 59.0 \metricpm{2.2} \\

      \textbf{\ppol{}: Evolved} & \textbf{0.825 \metricpm{0.285}} & \textbf{0.609 \metricpm{0.076}} & \textbf{0.717 \metricpm{0.205}} & \textbf{71.0 \metricpm{4.1}} & 58.1 \metricpm{1.3} & 62.2 \metricpm{5.5} & \textbf{49.4 \metricpm{8.5}} & \textbf{60.2 \metricpm{2.8}} \\

    \midrule
    \multicolumn{9}{l}{Dataset: \textbf{RealUserSim (WildChat)}, User Simulator: Qwen3-Next-80B-A3B-Instruct} \\
    \hline

    \rowcolor{lightgreen} Humans & 0.989 \metricpm{0.036} & 0.700 \metricpm{0.009} & 0.845 \metricpm{0.019} & 81.2 \metricpm{3.2} & 92.9 \metricpm{4.6} &  86.0 \metricpm{4.9} & 65.5 \metricpm{13.9} & 81.4 \metricpm{4.0} \\
    
    Base-simulator & 0.073 \metricpm{0.175} & 0.002 \metricpm{0.018} & 0.038 \metricpm{0.081} & 33.6 \metricpm{4.0} & \textbf{62.9 \metricpm{4.7}} & 36.7 \metricpm{2.7} & 15.9 \metricpm{1.7} & 37.3 \metricpm{1.8} \\
    
    DP Personas & 0.322 \metricpm{0.352} & 0.000 \metricpm{0.000} & 0.161 \metricpm{0.048} & 36.0 \metricpm{0.7} & 43.7 \metricpm{1.0} & 36.9 \metricpm{1.4} & 15.3 \metricpm{0.8} & 33.0 \metricpm{0.5} \\
    
    \ppol{}: Initial & 0.542 \metricpm{0.392} & 0.005 \metricpm{0.025} & 0.273 \metricpm{0.044} & 44.0 \metricpm{1.1} & 50.1 \metricpm{2.1} & \textbf{45.9 \metricpm{1.8}} & 21.5 \metricpm{1.1} & 40.4 \metricpm{0.9} \\
    
    \textbf{\ppol{}: Evolved} & \textbf{0.990 \metricpm{0.061}} & \textbf{0.533 \metricpm{0.160}} & \textbf{0.762 \metricpm{0.082}} & \textbf{62.8 \metricpm{3.2}} & 56.4 \metricpm{1.8} & 35.2 \metricpm{4.0} & \textbf{44.2 \metricpm{7.7}} & \textbf{49.7 \metricpm{2.3}} \\

    \bottomrule
  \end{tabular}
  }
\end{table*}

\vspace{-.2em}
Table~\ref{tab:persona_benchmark} shows that evolved \ppol{} consistently achieves the highest fitness score across all evaluated combinations of domains and user-simulator baselines. Relative to the Base-simulator, \ppol{} improves by \scoregain{+27.7 pp} on ColBench and \scoregain{+72.4 pp} on RealUserSim (WildChat). These gains effectively bridge a major portion of the quantitative gap between cooperative simulators and real human users. 
Appendix~\ref{app:pooled-results} includes results across 3 different user simulator LLMs in $\tau^2$ Retail and Airline domains.

\vspace{-.2em}
\subsection{Ablation Studies}
\vspace{-.2em}

\textbf{The role of evolutionary search.}
Simply prompting an LLM to generate personas (\emph{DP Personas}) or using our structured two-stage generator without evolution (\emph{\ppol{}: Initial}) yields only marginal gains over the base simulator, and both struggle to achieve high coverage. Evolution overcomes this limitation: as shown in Figure~\ref{fig:evolution-curves}, coverage starts lower for smaller $N$ early in the curriculum, then steadily improves as optimization scales to larger populations. Together, these results confirm that guided evolutionary search discovers a broader range of human-like behaviors than zero-shot generation. \looseness=-1

\textbf{Robustness to the behavioral representation.}
To test whether these improvements depend on the specific hand-designed features, we replace the 19-dimensional fingerprint throughout the pipeline, in both the human-likeness discriminator and the coverage objective, with general-purpose sentence embeddings. Repeating the full evolutionary loop attains comparable human-likeness and fitness on the held-out test split (HL $0.70$ vs.\ $0.78$; Score $0.63$ vs.\ $0.69$; Appendix~\ref{app:sbert}), showing that PPol can obtain similar gains under a substantially different behavioral representation.

\textbf{Robustness to the seed generator.}
We also examine whether the behaviors \ppol{} produces are an artifact of the specific diversity axes we seed in the program. Initializing the evolution from purely cooperative axes $D$ (polite, forthcoming, trusting, and patient) still recovers a broad range of human behavior, matching a contrastive seed run (HL $0.84$ vs.\ $0.80$; Coverage $0.70$ vs.\ $0.62$; Appendix~\ref{app:seed}). Friction is thus recovered from the data where warranted, rather than being introduced through the initialization. \looseness=-1

% \vspace{-.2em}
% \subsection{Ablation Studies}
% \vspace{-.2em}

% Our ablation baselines reveal that simply prompting an LLM to generate personas (\emph{DP Personas}) or using our structured two-stage generator without evolution (\emph{\ppol{}: Initial}) yields only marginal improvements over the base simulator. In particular, the unevolved baselines struggle to achieve high behavioral Coverage. Evolution overcomes this limitation: \emph{\ppol{}: Evolved} raises Coverage to levels nearing the human reference (e.g., $0.602$ vs.\ $0.614$ for Retail). As shown in the score trajectories (Figure~\ref{fig:evolution-curves}), coverage is naturally lower early in the curriculum for smaller $N$, but steadily improves as the optimization pressure shifts toward larger populations. This demonstrates that zero-shot prompting is insufficient for capturing the full spectrum of human behavior; guided search is essential.

\begin{wrapfigure}{r}{0.46\textwidth}
\centering
\vspace{-2em}
\includegraphics[width=\linewidth]{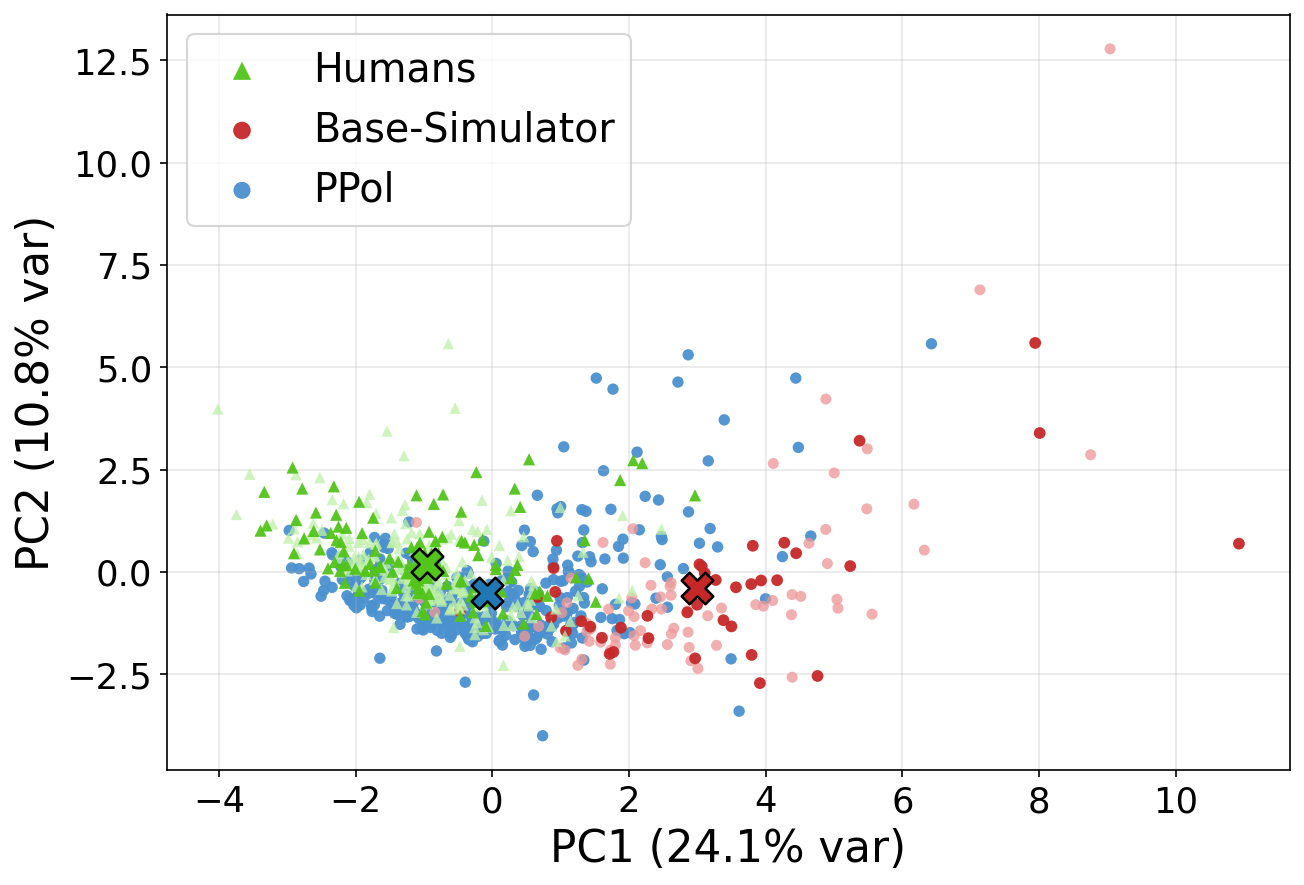}
\caption{PCA projection of Humans, \ppol{} and Base-Simulators in behavioral fingerprint-space. Domain: Retail, LLM: DeepSeek-V3.1.}
\vspace{-1em}
\label{fig:pca-scatterplot}
\end{wrapfigure}

\vspace{-.1em}
\subsection{Dimension-level Alignment}
Analyzing the S{\o}rensen--Dice alignment clarifies where these gains originate. While the base-simulator already exhibits moderate alignment on basic information disclosure ($D_2$), it fails on interactional dimensions like communication style ($D_1$) and error reaction ($D_4$). Evolved personas drive massive improvements in USI$_{D1-D4}$. For example, in the Retail domain, alignment nearly \textbf{doubles} on $D_3$ (from 35.7\% to 70.4\%) and more than \textbf{triples} on $D_4$ (from 23.3\% to 76.4\%). Visually, this shift is evident in the PCA projection of the fingerprint space (Figure~\ref{fig:pca-scatterplot}), where the evolved trajectories successfully expand from the narrow and not human-like base-simulator cluster to cover the broader human reference distribution. In Appendix~\ref{app:distr-coverage}, we report each population's mean and distribution Overlap with real humans (0 = no shared values, 1 = identical distributions). Overlap is high only when the simulator matches the human shape (both range and frequency), as under-covering or over-producing a region pulls it down. Equivalently, overlap = 1 $-$ total-variation distance. Averaged over all features, PPol's population \textbf{overlaps 79\% of the real-human distribution} vs the base simulator's 56\%.

\begin{wrapfigure}{r}{0.43\textwidth}
\centering
\vspace{-3em}
\includegraphics[width=\linewidth]{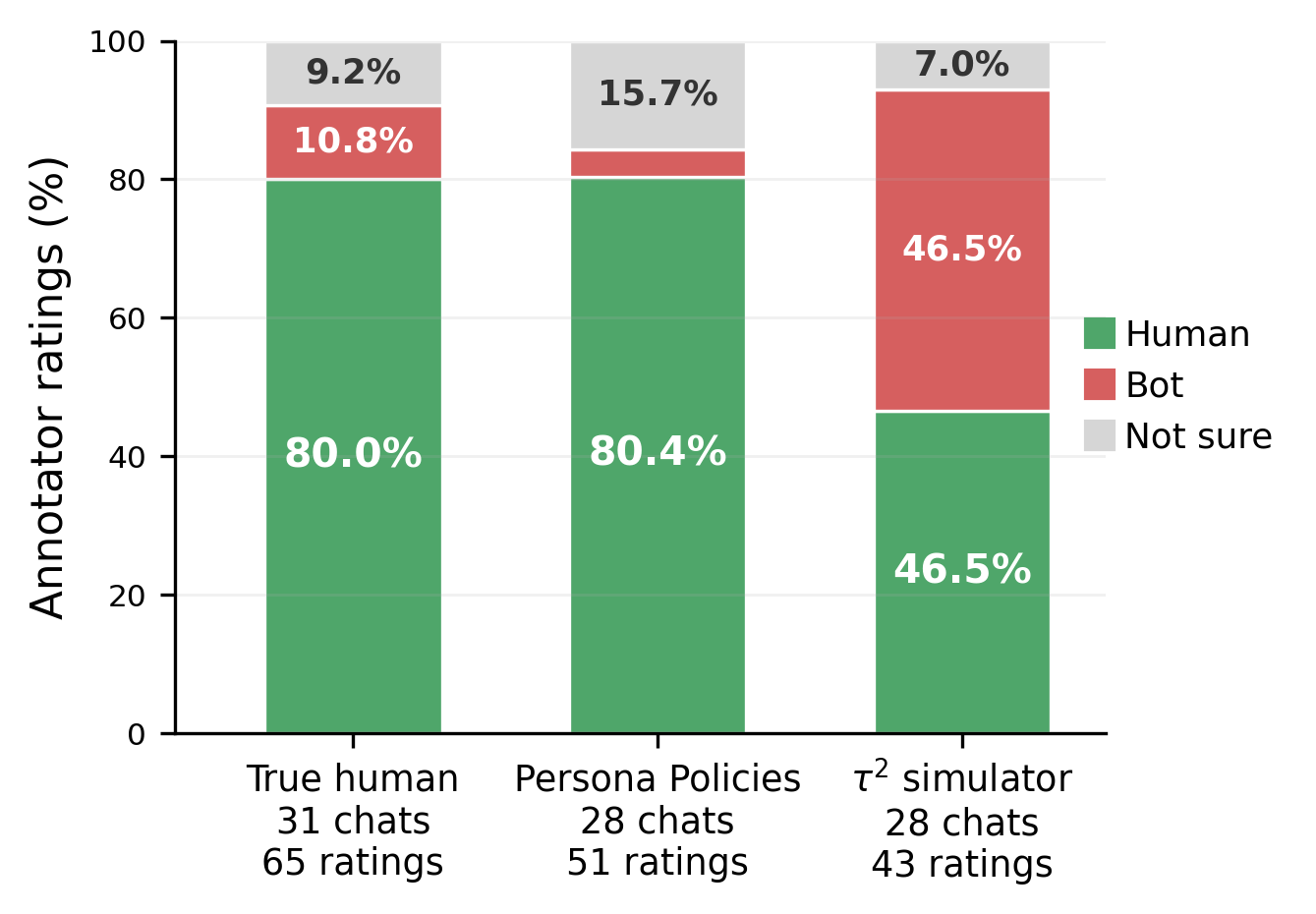}
\vspace{-1.4em}
\caption{Annotators judged \ppol{} utterances as \textit{Human} at the same rate as real human utterances, $\sim$2x more than baseline.}
\vspace{-2em}
\label{fig:human_eval_retail}
\end{wrapfigure}

\vspace{-.1em}
\subsection{Human Evaluation}
\label{sec:exp-human-eval}
To validate that gains in \textit{behavioral fingerprint} space correspond to real human likeness as judged by real humans, we ran a blinded evaluation on $\tau^2$-bench Retail conversations. We recruited 20 participants via ~\cite{prolific2026} under an Institutional Review Board approved protocol. Each transcript paired the same assistant-agent (Gemma-4-31B) with one of three user sources: a real human, the $\tau^2$ base-simulator, or our Persona Policy-conditioned simulator. Both simulators used DeepSeek-V3.1. 
Annotators were asked to judge whether the user side of the messages appeared to be a real human, a bot, or ambiguous. 
After quality filtering (see Appendix~\ref{app:human-eval}), the evaluation contains 16 annotators and 87 unique conversations.
Figure~\ref{fig:human_eval_retail} shows the results: \ppol{} were judged as \emph{`Human'} 80.4\% of the time, nearly the same rate as real human traces (80.0\%). In contrast, the $\tau^2$ simulator was identified as human only 46.5\% of the time, equal to the number of times it was recognized as a bot. Treating \emph{Human}=1 and \emph{Bot} or \emph{Not sure}=0, a Welch test confirmed significant difference with $t=3.556$ and $p=6.37{\times}10^{-4}$. 
% Excluding ``\emph{Not sure}'' responses strengthens the result: 95.3\% of \ppol{} are rated as human versus 50.0\% for the default simulator ($p=2.92{\times}10^{-6}$).
Additionally, our discriminator model's \textbf{$P(\mathrm{human})$ scores correlate strongly with human judgments} (Point-Biserial $r=0.49$, $p<0.001$), confirming that our 19-dimensional behavioral fingerprint serve as a grounded proxy for human judgment. 
\looseness=-1

\subsection{Qualitative Analysis}
\label{sec:exp-qualitative-generator}
\vspace{-.2em}

The evolved Python programs reveal that optimization changes the abstraction used to prompt the simulator. We seeded the initial generator with generic behaviors like ``terse'' or ``ambiguous.'' In contrast, successful evolved programs discover highly operational behavioral axes, such as \emph{information\_velocity}, \emph{temperament}, and \emph{assertive}, which influence the simulator to send fragmented messages, or push back against agent. 
Figure~\ref{fig:samples} shows samples of behaviors and personas. Rather than providing static character bios, \ppol{} leads to dynamic, turn-by-turn dialogue mechanics. For instance, depending on the task and active behavior axes, evolved prompts might explicitly instruct the simulator to ``\textit{not provide order numbers in the opening message,}'' ``\textit{answer only one requested field at a time,}'' or ``\textit{use lowercase and shorthand}'', which helps alter the conversational route. 

The behavioral coverage objective prevents convergence to generic (e.g., simply \textit{angry}) profiles. The evolved program constructs distinct situational contexts: a distracted commuter, a privacy-conscious skeptic, or a forgetful elder. Each persona anchors its disclosure and communication style in specific cognitive states, yielding varied trajectories for the same task. 
Reflections on rollout excerpts at each iteration reinforce successful behaviors and refine rigid or ineffective instructions, which lead to targeted code edits that push the simulators away from \textit{bot-like} behavior. \looseness=-1

\subsection{Training Agents with {}\ppol{}}
\label{sec:exp-agent-training}
\vspace{-.2em}

We study whether training agents on \ppol{}-generated conversations improves robustness to difficult, out-of-distribution users. We post-train a Gemma-4-31B agent with Supervised Fine-Tuning (SFT) on successful $\tau^2$-bench traces under two strictly matched regimes: (i) base-simulator trajectories (\emph{Base-only SFT}) and (ii) a mixture of base-simulator and \ppol{} trajectories (\ppol{}-\emph{augmented SFT}). Stress-testing these agents against four challenging $\tau$-trait simulators from \citet{he2025impatient} (Skeptical, Incoherent, Impatient, Confusion) shows that adding \ppol{} trajectories during SFT consistently outperforms \emph{Base-only} (\textcolor{darkgreen}{\textbf{$+$17\%}} relative gain on Retail and \textcolor{darkgreen}{\textbf{$+$8\%}} on Airline+Retail; Tables in Appendix~\ref{app:agent-transfer-extra}). Because both variants share identical hyperparameters, optimization steps, and data volume, these gains isolate the direct benefit of training on \ppol{}'s behavioral diversity. \looseness=-1

\begin{wraptable}{r}{0.45\textwidth}
\centering
\vspace{-1.2em}
\caption{\ppol{}-augmented SFT agent performs better in live human interactions.}
\resizebox{\linewidth}{!}{%
\setlength{\tabcolsep}{3pt}
\begin{tabular}{l >{\columncolor{basebg}}c >{\columncolor{ppolbg}}c c}
\toprule
Metric & Base & \ppol{} & $\Delta$ \\
\midrule
Task success ($\uparrow$)       & 29.5\% & 36.4\% & $+23.4\%$ \\
Perceived complete ($\uparrow$) & 56.0\% & 62.6\% & $+11.7\%$ \\
Rated easy ($\uparrow$)         & 39.7\% & 43.5\% & $+9.4\%$ \\
Unwanted actions ($\downarrow$) & 18.6\% & 10.2\% & $-45.2\%$ \\
Tool calls$/$chat ($\downarrow$) & 7.93  & 4.98  & $-37.2\%$ \\
Turns$/$chat ($\downarrow$)     & 13.5   & 15.2   & $+12.7\%$ \\
\bottomrule
\end{tabular}
}
\vspace{-.5em}
\label{tab:int-human-eval}
\end{wraptable}

\textbf{Interactive User Study.} To evaluate whether \ppol{}-augmented training improves performance in real-world, we designed a novel multi-turn interaction interface (Section~\ref{sec:experiments} \& Appendix~\ref{app:train-human-eval}) and recruited 36 participants on \citet{prolific2026}.
% Each participant completed 10 interactive scenarios in a randomized order: 5 chats were assigned to the agent trained on \ppol{} evolved personas, and 5 were assigned to the agent trained on the base-simulator. Participants were blinded to which agent they were interacting with and were instructed with the following guidelines. 
Across 239 live chats, \ppol{}-\textit{augmented} SFT agent outperformed the baseline across various study metrics (Table~\ref{tab:int-human-eval}). Beyond a boost in objective task success ($36.4\%$ vs.\ $29.5\%$), the model demonstrated substantial qualitative gains. Human ratings reported $11.7\%$ higher perceived task completion and $45.2\%$ fewer unwanted actions. It also eliminated redundant tool-calling behavior, sharply reducing tool-call variance ($\text{SD}{=}3.9$ down from $19.6$). Human ratings confirm that \ppol{} produces cleaner, more stable real-world dialogues. Task-stratified odds ratios demonstrate a clear preference for \ppol{}-augmented training, doubling the odds of task success ($\text{OR}{=}2.06$) while reducing unwanted actions ($\text{OR}{=}0.62$). \looseness=-1

\subsection*{Limitations}

\ppol{} relies on a corpus of real human dialogues in the target domain to train the discriminator and define the coverage reference. The method's application therefore depends on how accessible and representative the data is of the reference population. \ppol{} can only be as human-like as the data it is grounded in. We grounded our behavioral fingerprint vector in various language and communication style features, and showed that \ppol{} can work with latent space embeddings too. Our framework has scope to explore richer features that measure deeper aspects of human behavior such as internal states, memory, or intent. We see this as an important avenue for future work in developing better persona representations that are faithful to implicit user cues in continually evolving task settings. 

\vspace{-.2em}
\section{Conclusion}
\vspace{-.4em}

We introduced \textbf{Persona Policies} (\ppol{}), a plug-and-play control layer that induces realistic behaviors in user simulators without modifying original task goals. Our framework evolves persona-generating programs guided by a multi-objective score combining human-likeness and behavioral coverage, alongside natural-language reflection. The evolutionary process consistently discovers behavioral dimensions that default simulators lack. Across multiple domains and user-simulator backends, evolved \ppol{} improves the fitness score by $28$--$72\%$ points over the baseline. In blinded evaluations, annotators rated \ppol{}-generated conversations as `\emph{human}' 80.4\% of the time versus 46.5\% for the default simulator. In our live study, across 239 human-agent interactions, agents trained on \ppol{}-rollouts improved task success by 23.4\% and reduced unwanted actions by 45.2\% relative to baseline, proving more robust to real interactional friction. As agents move into complex social settings, user-simulators must be measured alongside agent performance so benchmarks remain meaningful proxies for reality. \looseness=-1

\section*{Acknowledgement}

This research was supported by the UW-Amazon Science Gift Hub, UW-Tsukuba Amazon NVIDIA Cross Pacific AI Initiative (XPAI), Tinker Research Grants, Open Philanthropy, Coefficient Giving, Toyota Research Institute Kaizen Initiative, the Schmidt AI2050 Fellows program, and the NSF CISE RI program, award \#2550849. This work was supported in part by Advanced Micro Devices, Inc. under the AMD University Program’s AI \& HPC Cluster. This material is based upon work supported by the Defense Advanced Research Projects Agency and the Air Force Research Laboratory, contract number(s): FA8650-23-C-7316. 
This work is also supported by the U.S. National Science Foundation (NSF) Award IIS-2336769.
This work was also supported in part by the U.S. National Science Foundation (NSF) CAREER Award 2337877, Schmidt Sciences Award on AI \& Advanced Computing, through the Science of Trustworthy AI program, and by the University of Washington Tech Policy Lab. This work was also supported in part by the Robert L. McDevitt, K.S.G., K.C.H.S.\ and Catherine H. McDevitt L.C.H.S.\ Chair in Computer Science at Georgetown University.
Any opinions, findings and conclusions, or recommendations expressed in this material are those of the author(s) and do not necessarily reflect the views of AFRL, DARPA, NSF or Schmidt Sciences.

{
\bibliographystyle{conference} % NeurIPS often uses plainnat or unsrtnat
\bibliography{refs}  % Points to your references.bib file
}

\section*{Ethics Statement}
All human evaluation studies were conducted under IRB approval. Annotators were fairly compensated, briefed on data usage, and provided informed consent. The user simulator framework and behavioral datasets developed in this work do not contain personally identifiable information (PII) and are designed to improve agent robustness under realistic human interactions.
Our framework generates realistic, diverse user behaviors to evaluate and train task-oriented agents, helping prevent brittle failures in deployments like customer support and enabling proactive safety and fairness auditing. While realistic user simulation carries potential dual-use risks, such as making deceptive systems more convincing, our evaluation is strictly constrained to benign benchmark environments with fixed goals and safety boundaries. To eliminate misuse risk in other settings, future work should pair persona generation with explicit safeguards, including task screening, safety filters, and non-toxicity and fairness constraints in the evolutionary objectives. These safeguards can help ensure that realistic simulation improves agent robustness without amplifying harmful human biases or enabling unsafe tasks. \looseness=-1

%%%%%%%%%%%%%%%%%%%%%%%%%%%%%%%%%%%%%%%%%%%%%%%%%%%%%%%%%%%%%%%%%%%%%%%%%%%%%%%%%%%%%%%%%%%%%%%%%%%%%%%%%%%%%%%%%%%%%%%%%%%%%%%%%%%%%%%%%%%%%%%%%%%%%%%%%%%%%%%%%%%%%%%%%%%%%%%%%%%%

%%%%%%%%%%%%%%%%%%%%%%%%%%%%%%%%%%%%%%%%%%%%%%%%%%%%%%%%%%%%

\newpage
\appendix

\section*{Technical Appendices}
\label{appendix}

\section{Full Results: \texorpdfstring{$\tau^2$}{tau-2}-bench Retail and Airline}
\label{app:pooled-results}

\begin{table*}[ht]
  \centering
  \scriptsize
  \caption{\textbf{Retail domain ($\tau^2$-bench).}
  HL is the mean discriminator probability $\overline{P}(\mathrm{human})$, Coverage is the mean behavioral coverage $\overline{B}_{\mathrm{cover}}$, and Score is the fitness $\mathcal{M}$ from Eq.\eqref{eq:combined_score} reported on the $[0,1]$ scale.
  $D_1$--$D_4$ are S{\o}rensen--Dice alignment scores for communication style, information disclosure, clarification behavior, and error reaction, respectively. Values are reported as (mean$\pm{\text{std}}$).
  % Metrics are computed with N=3 annotators per task for Humans, and N=10 personas per task for Persona Policies (\ppol{}).
  }
  \vspace{1em}
  \label{tab:persona_benchmark_retail_full}
  \resizebox{\linewidth}{!}{%
  \begin{tabular}{l c c c | c c c c c}
    \toprule
    & \multicolumn{3}{c|}{Optimization Metrics $\uparrow$} & \multicolumn{5}{c}{Behavioral Alignment (\%) $\uparrow$} \\
    \cmidrule(lr){2-4} \cmidrule(lr){5-9}
    Method & HL & Coverage  & Score  & $D_1$  & $D_2$  & $D_3$  & $D_4$  & $\mathrm{USI}_{\mathrm{D1-D4}}$ \\
    
    \midrule
    \multicolumn{9}{l}{User Simulator: DeepSeek-V3.1} \\
    \hline
    
    \rowcolor{lightgreen} Humans  & 0.956 \metricpm{0.076} & 0.614 \metricpm{0.091} & 0.785 \metricpm{0.051} & 84.3 \metricpm{2.4} & 97.6 \metricpm{2.2} & 87.2 \metricpm{6.2} & 82.2 \metricpm{12.8} & 87.8 \metricpm{3.5} \\
    
    Base-simulator & 0.112 \metricpm{0.158} & 0.132 \metricpm{0.177} & 0.122 \metricpm{0.147} & 28.8 \metricpm{2.9} & 80.0 \metricpm{4.9} & 51.5 \metricpm{6.9} & 42.8 \metricpm{9.2} & 50.8 \metricpm{2.9} \\

    DP Personas & 0.283 \metricpm{0.313} & 0.261 \metricpm{0.274} & 0.272 \metricpm{0.223} & 35.7 \metricpm{1.3} & 80.7 \metricpm{2.7} & 53.1 \metricpm{2.6} & 30.9 \metricpm{3.0} & 50.1 \metricpm{1.6} \\

    \ppol{}: Initial & 0.394 \metricpm{0.314} & 0.213 \metricpm{0.220} & 0.303 \metricpm{0.148} & 40.6 \metricpm{1.9} & \textbf{87.8 \metricpm{2.4}} & 50.0 \metricpm{2.5} & 28.9 \metricpm{2.4} & 51.8 \metricpm{1.6} \\

	    \textbf{\ppol{}: Evolved} & \textbf{0.657 \metricpm{0.266}} & \textbf{0.608 \metricpm{0.144}} & \textbf{0.633 \metricpm{0.123}} & \textbf{46.8 \metricpm{2.4}} & 82.4 \metricpm{3.7} & \textbf{84.0 \metricpm{4.8}} & \textbf{60.1 \metricpm{8.3}} & \textbf{68.3 \metricpm{2.6}} \\

    \midrule
    \multicolumn{9}{l}{User Simulator: GPT-5.4-Mini} \\
    \hline

    \rowcolor{lightgreen} Humans & 0.916 \metricpm{0.131} & 0.614 \metricpm{0.091} & 0.765 \metricpm{0.063} & 84.3 \metricpm{2.5} & 97.6 \metricpm{1.8} & 87.2 \metricpm{6.1} & 82.2 \metricpm{12.5} & 87.8 \metricpm{3.5} \\

    Base-simulator & 0.212 \metricpm{0.223} & 0.102 \metricpm{0.174} & 0.157 \metricpm{0.171} & 34.6 \metricpm{2.5} & 72.0 \metricpm{4.8} & 40.0 \metricpm{6.6} & 42.4 \metricpm{8.2} & 47.2 \metricpm{3.3} \\

    DP Personas & 0.283 \metricpm{0.280} & 0.197 \metricpm{0.265} & 0.240 \metricpm{0.204} & 41.6 \metricpm{2.4} & 65.5 \metricpm{4.1} & 50.9 \metricpm{3.7} & 36.8 \metricpm{4.9} & 48.7 \metricpm{2.8} \\

    \ppol{}: Initial & 0.309 \metricpm{0.273} & 0.144 \metricpm{0.217} & 0.227 \metricpm{0.142} & 43.1 \metricpm{2.8} & 71.1 \metricpm{3.5} & 53.3 \metricpm{3.9} & 35.6 \metricpm{4.2} & 50.8 \metricpm{2.1} \\

    \textbf{\ppol{}: Evolved} & \textbf{0.623 \metricpm{0.326}} & \textbf{0.548 \metricpm{0.182}} & \textbf{0.586 \metricpm{0.148}} & \textbf{63.2 \metricpm{3.2}} & \textbf{90.1 \metricpm{2.8}} & \textbf{87.1 \metricpm{5.3}} & \textbf{58.5 \metricpm{6.8}} & \textbf{74.7 \metricpm{2.4}} \\
    
    \midrule
    \multicolumn{9}{l}{User Simulator: Qwen3-Next-80B-A3B-Instruct} \\
    \hline

    \rowcolor{lightgreen} Humans & 0.953 \metricpm{0.076} & 0.614 \metricpm{0.091} & 0.783 \metricpm{0.055} & 84.3 \metricpm{2.3} & 97.6 \metricpm{2.2} & 87.2 \metricpm{6.0} & 82.2 \metricpm{12.6} & 87.8 \metricpm{3.5} \\

    Base-simulator & 0.107 \metricpm{0.244} & 0.046 \metricpm{0.136} & 0.077 \metricpm{0.186} & 24.0 \metricpm{2.4} & 57.5 \metricpm{5.8} & 35.7 \metricpm{6.3} & 23.3 \metricpm{6.6} & 35.1 \metricpm{2.9} \\

    DP Personas & 0.291 \metricpm{0.355} & 0.100 \metricpm{0.219} & 0.196 \metricpm{0.212} & 29.5 \metricpm{1.8} & 56.5 \metricpm{2.8} & 37.8 \metricpm{3.5} & 25.2 \metricpm{3.1} & 37.2 \metricpm{2.0} \\

    \ppol{}: Initial & 0.356 \metricpm{0.348} & 0.017 \metricpm{0.085} & 0.186 \metricpm{0.083} & 31.4 \metricpm{2.1} & 58.7 \metricpm{2.7} & 38.7 \metricpm{2.3} & 30.3 \metricpm{2.4} & 39.8 \metricpm{1.4} \\

    \textbf{\ppol{}: Evolved} & \textbf{0.784 \metricpm{0.204}} & \textbf{0.602 \metricpm{0.161}} & \textbf{0.693 \metricpm{0.139}} & \textbf{69.6 \metricpm{5.0}} & \textbf{89.8 \metricpm{2.4}} & \textbf{70.4 \metricpm{4.5}} & \textbf{76.4 \metricpm{8.1}} & \textbf{76.5 \metricpm{2.6}} \\

    \bottomrule
  \end{tabular}
  }
\end{table*}

\begin{table*}[ht]
  \centering
  \scriptsize
  \caption{\textbf{Airline domain ($\tau^2$-bench).}
  Column definitions match Table~\ref{tab:persona_benchmark_retail_full}}
  \vspace{1em}
  \label{tab:persona_benchmark_airline}
  \resizebox{\linewidth}{!}{%
  \begin{tabular}{l c c c | c c c c c}
    \toprule
    & \multicolumn{3}{c|}{Optimization Metrics $\uparrow$} & \multicolumn{5}{c}{Behavioral Alignment (\%) $\uparrow$} \\
    \cmidrule(lr){2-4} \cmidrule(lr){5-9}
    Method & HL & Coverage  & Score  & $D_1$  & $D_2$  & $D_3$  & $D_4$  & $\mathrm{USI}_{\mathrm{D1-D4}}$ \\
    \midrule
    \multicolumn{9}{l}{User Simulator: DeepSeek-V3.1} \\
    \hline

    \rowcolor{lightgreen} Humans & 0.932 \metricpm{0.109} & 0.584 \metricpm{0.103} & 0.758 \metricpm{0.062} & 91.3 \metricpm{3.9} & 98.1 \metricpm{2.0} & 84.8 \metricpm{9.1} & 77.8 \metricpm{11.2} & 88.0 \metricpm{3.8} \\

    Base-simulator & 0.173 \metricpm{0.152} & 0.112 \metricpm{0.180} & 0.143 \metricpm{0.153} & 38.8 \metricpm{4.3} & 83.6 \metricpm{4.7} & 56.9 \metricpm{9.1} & 45.4 \metricpm{8.2} & 56.2 \metricpm{3.5} \\

    DP Personas & 0.353 \metricpm{0.294} & 0.260 \metricpm{0.222} & 0.306 \metricpm{0.144} & 36.1 \metricpm{1.5} & 71.9 \metricpm{1.9} & 43.6 \metricpm{3.0} & 40.8 \metricpm{3.6} & 48.1 \metricpm{1.6} \\

    \ppol{}: Initial & 0.466 \metricpm{0.331} & 0.264 \metricpm{0.217} & 0.365 \metricpm{0.131} & 49.1 \metricpm{3.2} & 80.4 \metricpm{2.0} & 44.9 \metricpm{3.5} & 46.2 \metricpm{3.5} & 55.2 \metricpm{2.1} \\

    \textbf{\ppol{}: Evolved} & \textbf{0.534 \metricpm{0.290}} & \textbf{0.613 \metricpm{0.102}} & \textbf{0.574 \metricpm{0.120}} & \textbf{57.0 \metricpm{2.3}} & \textbf{91.8 \metricpm{1.4}} & \textbf{69.3 \metricpm{3.9}} & \textbf{65.2 \metricpm{8.1}} & \textbf{70.8 \metricpm{2.4}} \\

    \midrule
    \multicolumn{9}{l}{User Simulator: GPT-5.4-Mini} \\
    \hline

    \rowcolor{lightgreen} Humans & 0.903 \metricpm{0.103} & 0.584 \metricpm{0.103} & 0.744 \metricpm{0.067} & 91.3 \metricpm{4.0} & 98.1 \metricpm{2.0} & 84.8 \metricpm{8.9} & 77.8 \metricpm{11.5} & 88.0 \metricpm{3.8} \\

    Base-simulator & 0.322 \metricpm{0.257} & 0.163 \metricpm{0.220} & 0.243 \metricpm{0.222} & 47.2 \metricpm{6.2} & 62.3 \metricpm{6.4} & 61.1 \metricpm{11.1} & 46.8 \metricpm{11.5} & 54.3 \metricpm{5.2} \\

    DP Personas & 0.358 \metricpm{0.295} & 0.196 \metricpm{0.218} & 0.277 \metricpm{0.182} & 44.5 \metricpm{3.2} & 59.4 \metricpm{4.2} & 62.6 \metricpm{6.5} & 42.9 \metricpm{5.7} & 52.4 \metricpm{2.9} \\

    \ppol{}: Initial & 0.457 \metricpm{0.308} & 0.164 \metricpm{0.230} & 0.310 \metricpm{0.144} & 59.3 \metricpm{4.3} & 65.6 \metricpm{4.2} & 57.6 \metricpm{7.0} & 54.4 \metricpm{5.4} & 59.2 \metricpm{3.3} \\

    \textbf{\ppol{}: Evolved} & \textbf{0.604 \metricpm{0.320}} & \textbf{0.545 \metricpm{0.142}} & \textbf{0.574 \metricpm{0.120}} & \textbf{68.7 \metricpm{4.5}} & \textbf{87.7 \metricpm{3.9}} & \textbf{66.9 \metricpm{8.3}} & \textbf{66.7 \metricpm{8.2}} & \textbf{72.5 \metricpm{3.6}} \\
    
    \midrule
    \multicolumn{9}{l}{User Simulator: Qwen3-Next-80B-A3B-Instruct} \\
    \hline

    \rowcolor{lightgreen} Humans & 0.962 \metricpm{0.061} & 0.584 \metricpm{0.103} & 0.773 \metricpm{0.067} & 91.3 \metricpm{3.8} & 98.1 \metricpm{2.1} & 84.8 \metricpm{8.9} & 77.8 \metricpm{11.4} & 88.0 \metricpm{3.8} \\

    Base-simulator & 0.128 \metricpm{0.154} & 0.000 \metricpm{0.000} & 0.064 \metricpm{0.077} & 22.4 \metricpm{1.8} & 49.6 \metricpm{3.7} & 29.0 \metricpm{6.0} & 63.4 \metricpm{12.1} & 41.1 \metricpm{3.8} \\

    DP Personas & 0.337 \metricpm{0.325} & 0.051 \metricpm{0.159} & 0.194 \metricpm{0.163} & 28.9 \metricpm{2.1} & 52.0 \metricpm{2.9} & 39.4 \metricpm{5.6} & 43.3 \metricpm{3.7} & 40.9 \metricpm{2.2} \\

    \ppol{}: Initial & 0.468 \metricpm{0.353} & 0.011 \metricpm{0.045} & 0.239 \metricpm{0.066} & 35.0 \metricpm{2.2} & 57.1 \metricpm{3.2} & 43.7 \metricpm{3.7} & 48.8 \metricpm{4.2} & 46.2 \metricpm{1.6} \\

    \textbf{\ppol{}: Evolved} & \textbf{0.761 \metricpm{0.280}} & \textbf{0.483 \metricpm{0.184}} & \textbf{0.622 \metricpm{0.135}} & \textbf{54.7 \metricpm{4.2}} & \textbf{92.5 \metricpm{2.8}} & \textbf{80.0 \metricpm{5.3}} & 62.5 \metricpm{4.9} & \textbf{72.4 \metricpm{2.8}} \\

    \bottomrule
  \end{tabular}
  }
\end{table*}

\section{Dataset Statistics}
\label{app:datasets}

Each dataset provides tasks that are supposed to be completed in a user--agent multi-turn dialog. \ppol{} requires real-human dialogues in the targeted domain to serve as the human behavior reference distribution, from which the full set of behavioral fingerprint vectors $\mathcal{H}_{\mathrm{ref}}$ is derived. We partition the dialogues disjointly into a \textbf{train} portion, used to train the human-vs-simulator discriminator and as the coverage target during evolution, and a held-out \textbf{test} portion, used for the main evaluations and method comparisons (as the \textbf{Humans} reference row). The counts below give the size of each portion.

\paragraph{$\tau^2$-bench (Retail, Airline).} We use the official per-domain train/test partitions ($74/40$ tasks for Retail, and $30/20$ tasks for Airline), with a further $80:20$ train/validation hold-out inside the train split during evolution (Retail: $60/14$ tasks, Airline: $24/6$ tasks). Human behavior comes from the real user trajectories of \citet{zhou2026sim2real}, which provide three annotator dialogues per task, giving $222/120$ train/test human dialogues for Retail and $90/60$ for Airline.

\paragraph{ColBench.} Tasks are backend-programming scenarios~\citep{zhou2025sweet} ($10$k train and $5$k test available; we sample $1$k train and $100$ test). Human behavior comes from SWE-Chat~\citep{baumann2026swechat} developer--agent dialogues, filtered for English and code-authoring intent, yielding $1{,}449$ dialogues across $63$ repositories, split repository-disjointly into $1{,}088$ train ($34$ repos) and $361$ held-out ($29$ repos).

\paragraph{WildChat.} We use RealUserSim~\citep{zhu2026realusersim}: $600$ real WildChat~\citep{zhao2024wildchat} user--AI conversations spanning $460$ users and $72$ domains, each paired with an extracted problem description. Conversations are split user-disjointly into $450$ train and $150$ held-out tasks, with $446$ / $148$ human dialogues forming the train / test reference. Each task is a single real-user conversation.

\section{Human Behavior Reference}
\label{app:human-coverage}

For simulator methods, Coverage is computed between the group of personas generated for each task and the dataset/domain's human reference cloud $\mathcal{H}_{\mathrm{ref}}$, then averaged across tasks~(Eq.~\ref{eq:b_cover_task}). The \textbf{Humans} row is intended to provide an analogous estimate: on average, how broadly does a small group of real users performing the same (or similar) task span the wider range of human behavior observed across the domain? We therefore compute Coverage over task-level groups rather than comparing the complete human dataset with itself. Because each group contains only a small subset of users while $\mathcal{H}_{\mathrm{ref}}$ represents behavior across the full domain, Human Coverage is not expected to equal 1. In our $\tau^2$-bench's human reference data, each task has three real human dialogues~\citep{zhou2026sim2real}; we treat their fingerprints as one group, compute its Coverage against $\mathcal{H}_{\mathrm{ref}}$, and average across tasks. ColBench and WildChat contain only one real human dialogue per task, so a per-task group does not exist; we instead cluster semantically similar tasks and subsample human dialogues within each cluster as follows.

We embed the task descriptions ($c_t$) with TF--IDF and cluster it with $K$-means ($K=\lfloor n/12 \rfloor$); clusters smaller than 10 are merged into their nearest centroid, enforcing a minimum cluster size of 10. From each cluster we then draw $N{=}10$-sized subsamples of human dialogues (once if the cluster has exactly 10 members, and 100 times if it is larger), compute each subsample's Coverage against $\mathcal{H}_{\mathrm{ref}}$ (Eq.~\ref{eq:b_cover_task}), and report the mean over all subsamples. As in the main metric, fingerprints are $z$-scored by the human reference before Chamfer distances are computed, so that no single high-variance feature (e.g., message length) dominates, and $\mathcal{H}_{\mathrm{ref}}$ is the same human reference cloud used to score the simulator methods.

\section{Behavioral Fingerprint Features}
\label{app:fingerprint-features}

To compute the behavioral fingerprint for a user trajectory, we extract 19 scalar features grouped into four dimensions of human communication, inspired by the Sim2Real taxonomy \citep{zhou2026sim2real}. These features are computed strictly from the user's turns (not the agent's) using regular-expression matching and basic turn statistics. The full list of features is as follows:

\paragraph{D1: Communication Style (8 features).} Captures how the user talks:
\begin{itemize}
    \item \texttt{words\_per\_turn}: Average number of words per user message.
    \item \texttt{short\_utterance\_rate}: Fraction of turns that are extremely brief (e.g., $\le$ 3 words).
    \item \texttt{politeness\_rate}: Frequency of polite markers (e.g., ``please'', ``thank you'', ``appreciate'').
    \item \texttt{formality\_rate}: Frequency of formal vs.\ casual linguistic markers (e.g., ``moreover'', ``however'', ``regarding'').
    \item \texttt{acknowledgment\_rate}: Frequency of explicit short acknowledgments (e.g., ``ok'', ``got it'', ``sounds good'', ``understood'').
    \item \texttt{verbosity\_cv}: Coefficient of variation of turn lengths (capturing burstiness).
    \item \texttt{repetition\_rate}: How often the user repeats identical or highly overlapping phrases across turns.
    \item \texttt{identity\_confusion\_rate}: Instances where the user uses incorrect terminology or adopts agent-side phrasing (e.g., ``how may I help'', ``let me check'').
\end{itemize}

\paragraph{D2: Information Disclosure (3 features).} Captures how the user provides required information:
\begin{itemize}
    \item \texttt{front\_loading\_ratio}: The proportion of task-critical identifiers (order numbers, flight dates) provided in the first turn vs.\ later turns.
    \item \texttt{identifiers\_per\_turn}: Average number of entities provided per turn.
    \item \texttt{opening\_length}: Word count of the user's very first message.
\end{itemize}

\paragraph{D3: Clarification Behavior (5 features).} Captures how the user handles ambiguity:
\begin{itemize}
    \item \texttt{uncertainty\_rate}: Frequency of hesitant language (e.g., ``maybe'', ``I think'', ``not sure'', ``probably'').
    \item \texttt{certainty\_rate}: Frequency of definitive language (e.g., ``definitely'', ``absolutely'', ``for sure'', ``100\%'').
    \item \texttt{pushback\_rate}: Frequency of the user explicitly rejecting an agent's statement (e.g., ``that's not right'', ``I already told you'', ``you're not listening'').
    \item \texttt{clarification\_question\_rate}: Frequency of asking the agent to explain a term or step (e.g., ``what do you mean'', ``can you clarify'').
    \item \texttt{info\_seeking\_rate}: General rate of questions asked by the user (e.g., ``what is the status'', ``how do I'', ``when will'').
\end{itemize}

\paragraph{D4: Error Reaction (3 features).} Captures how the user responds to friction or agent mistakes:
\begin{itemize}
    \item \texttt{emotional\_expression\_rate}: Frequency of emotional markers (e.g., ``frustrated'', ``annoying'', ``ugh'', ``ridiculous'').
    \item \texttt{accusatory\_rate}: Frequency of placing blame or strong dissatisfaction (e.g., ``useless'', ``unacceptable'', ``scam'', ``worst'').
    \item \texttt{strategy\_pivot\_rate}: Instances where the user abruptly abandons one line of inquiry to try another (e.g., ``instead'', ``on second thought'', ``let's try'', ``scratch that'').
\end{itemize}

\section{Behavioral Discriminator Details}
\label{app:discriminator-details}

To evaluate human-likeness and compute the fitness signal during evolution, we train a Random Forest discriminator on behavioral fingerprints to distinguish real human dialogues from trajectories produced by the default $\tau^2$-bench user simulator. 

\begin{table}[h]
  \centering
  \small
  \caption{\textbf{Held-out Test Performance of Behavioral Discriminators.}}
  \label{tab:discriminator-performance}
  \begin{tabular}{l l c c c}
    \toprule
    \textbf{User Simulator Model} & \textbf{Domain} & \textbf{ROC-AUC} & \textbf{Accuracy} & \textbf{F1 Score} \\
    \midrule
    \multirow{2}{*}{DeepSeek-V3.1} & Retail & 1.000 & 0.988 & 0.992 \\
    & Airline & 0.998 & 0.975 & 0.983 \\
    \midrule
    \multirow{2}{*}{GPT-5.4-Mini} & Retail & 0.975 & 0.962 & 0.975 \\
    & Airline & 0.940 & 0.924 & 0.952 \\
    \midrule
    \multirow{2}{*}{Qwen3-Next-80B-A3B} & Retail & 0.982 & 0.981 & 0.988 \\
    & Airline & 0.990 & 0.975 & 0.984 \\
    \bottomrule
  \end{tabular}
\end{table}

\paragraph{Hyperparameters.}
For all domains, the discriminator is a \texttt{RandomForestClassifier} implemented via scikit-learn. The model is trained on the standardized 19-dimensional behavioral fingerprints using $200$ estimators, a maximum depth of $12$, and \texttt{class\_weight=``balanced"} to account for unequal numbers of human and simulator dialogues in the training splits. A consistent \texttt{random\_state=42} is used.

\begin{wraptable}{r}{0.4\textwidth}
  \centering
  \small
  \vspace{-1em}
  \caption{\textbf{Top Discriminative Features (Retail, DeepSeek-V3.1).} Gini importance scores from the fitted Random Forest classifier.}
  \label{tab:feature-importance}
  \begin{tabular}{l c}
    \toprule
    \textbf{Feature Name} & \textbf{Importance} \\
    \midrule
    \texttt{short\_utterance\_rate} & 0.257 \\
    \texttt{words\_per\_turn} & 0.240 \\
    \texttt{verbosity\_cv} & 0.165 \\
    \texttt{politeness\_rate} & 0.089 \\
    \texttt{acknowledgment\_rate} & 0.084 \\
    \texttt{formality\_rate} & 0.041 \\
    \texttt{opening\_length} & 0.034 \\
    \texttt{front\_loading\_ratio} & 0.024 \\
    \texttt{uncertainty\_rate} & 0.017 \\
    \texttt{identifiers\_per\_turn} & 0.016 \\
    \bottomrule
  \end{tabular}
  \vspace{-2em}
\end{wraptable}

\paragraph{Discriminator Performance.}
Because the base behavior of the user simulator depends heavily on its underlying LLM, we train a separate discriminator for each combination of domain and simulator model. The models are fit solely on dialogues corresponding to the official $\tau^2$-bench \textbf{train} tasks and evaluated on the held-out \textbf{test} tasks. Table~\ref{tab:discriminator-performance} summarizes the held-out test performance of the discriminators across all evaluated simulator variants. The high AUC and F1 scores indicate that the 19-dimensional behavioral fingerprint captures robust, discernible differences between real human users and the cooperative default simulators, regardless of the underlying model.

\paragraph{Feature Importance.}
To understand which behavioral dimensions most strongly distinguish real humans from the base-simulator, we analyze the Random Forest feature importances. Table~\ref{tab:feature-importance} lists the most discriminative features for the DeepSeek-V3.1 model on the Retail domain. Features tied to communication style (e.g., short utterance rate, verbosity variance) and pacing dominate the signal, confirming that the base-simulator struggles to replicate the structural and temporal nuances of real human typing.

\section{Distributional Overlap}
\label{app:distr-coverage}
Table~\ref{tab:distr_cov} presents six illustrative features spanning the four behavioral dimensions.

\begin{table*}[ht]
\centering
\begin{small}
\caption{How much of the human distribution does each simulator reproduce? Comparison of illustrative behavioral features across dimensions between Humans, \ppol{}, and the Base-simulator.}
\label{tab:distr_cov}
\begin{tabular}{lcccccc}
\toprule
\textbf{Feature} & \textbf{Human} & \textbf{\ppol{}} & \textbf{Base-simulator} & \multicolumn{2}{c}{\textbf{Overlap with Humans}} \\
\cmidrule{5-6}
& & & & \textbf{\ppol{}} & \textbf{Base-simulator} \\
\midrule
words\_per\_turn & 9.2 & 11.4 & 54.6 & 0.86 & 0.16 \\
formality\_rate & 0.02 & 0.05 & 0.75 & 0.92 & 0.05 \\
politeness\_rate & 0.15 & 0.14 & 0.50 & 0.69 & 0.40 \\
opening\_length & 20 & 13 & 72 & 0.53 & 0.21 \\
uncertainty\_rate & 0.05 & 0.04 & 0.24 & 0.76 & 0.53 \\
emotional\_expression\_rate & 0.008 & 0.012 & 0.100 & 0.92 & 0.72 \\
\midrule
\textit{All 19 features (average)} & & & & \textbf{0.79} & \textbf{0.56} \\
\bottomrule
\end{tabular}
\end{small}
\end{table*}

Averaged over all features, \ppol{}'s population overlaps $79\%$ of the real-human distribution vs the base simulator's $56\%$. For instance, \texttt{formality\_rate} of base-simulators in $\tau^2$-bench is found to be very high ($0.72$), whereas \ppol{} behaves much more similarly to humans (\ppol{}=$0.05$, humans=$0.02$), showing a large improvement in resemblance; specifically, \ppol{}'s gap to the human value is $23\times$ smaller than the base simulator's.

\section{Search and Compute Details}
\label{app:compute-details}

Evolution uses only the official $\tau^2$-bench train split, with a held-out validation slice for checkpoint selection; final benchmark numbers are reported on the official test split. Each fitness evaluation samples five training tasks and runs one episode for every generated persona on each task. We use a curriculum over the number of personas per task, $N=5 \rightarrow 8 \rightarrow 10$, so a candidate generator is evaluated with 25, 40, or 50 $\tau^2$ episodes per fitness call. Rollouts are parallelized with up to 30 workers, while candidate programs are evaluated sequentially. We checkpoint every iteration and select the final evolved generator by validation score averaged over $N \in \{5,8,10\}$.

The evolutionary archive uses a population size of 50, five islands, migration interval 5, migration rate 0.2, and an elite selection ratio of 0.2. The OpenEvolve evaluator timeout is 3600 seconds per full fitness evaluation, and generator/reflection calls use Gemini~3 Flash through LiteLLM with a direct Gemini fallback. At test time, all persona-based conditions use $N=10$ personas per task: up to 400 rollouts per retail condition and 200 rollouts per airline condition, compared with one rollout per task for the base-simulator.

\section{Ablation Studies}

\subsection{Changing Discriminator Feature Space}
\label{app:sbert}

Both our human-likeness and coverage signals operate on the 19-dimensional behavioral fingerprint. To test whether the gains from evolutionary search are specific to these hand-designed features, we replace the fingerprint \emph{everywhere} in the pipeline with general-purpose sentence embeddings and re-run the entire method end-to-end.

\paragraph{Setup.} Each trajectory is represented by the mean of its user-turn sentence embeddings (OpenAI \texttt{text-embedding-3-small}), which share none of the 19 features. The human-likeness discriminator becomes a logistic-regression classifier on these embeddings (human vs.\ base-simulator), and coverage is a Chamfer distance in a PCA-reduced embedding space. In this semantic space the two-sided Chamfer saturates, because the embeddings separate real humans from \emph{every} LLM simulator so sharply that no simulator overlaps the human cloud; we therefore use the one-sided (human-coverage) term, which is the quantity the search optimizes. We then run the full evolutionary search ($70$ iterations, $N{=}5{\to}8{\to}10$) under this embedding objective, select the best generator on validation, and evaluate on the held-out Retail test split.

\begin{table}[h]
  \centering
  \small
  \caption{\textbf{Feature-agnosticism on $\tau^2$-bench Retail (Qwen3-Next-80B-A3B), test split.} Replacing the 19-feature fingerprint everywhere with learned sentence embeddings yields nearly identical human-likeness and fitness. HL is mean $\overline{P}(\mathrm{human})$; Score is the fitness $\mathcal{M}$.}
  \label{tab:sbert-ablation}
  \begin{tabular}{l c c c}
    \toprule
    \textbf{Method (representation)} & \textbf{HL} & \textbf{Coverage} & \textbf{Score} \\
    \midrule
    Base-simulator (19-feat fingerprint) & 0.11 & 0.05 & 0.08 \\
    \textbf{\ppol{}: Evolved (19-feat fingerprint)} & \textbf{0.78} & \textbf{0.60} & \textbf{0.69} \\
    \midrule
    Base-simulator (learned embeddings) & 0.01 & 0.46 & 0.23 \\
    \textbf{\ppol{}: Evolved (learned embeddings)} & \textbf{0.70} & \textbf{0.56} & \textbf{0.63} \\
    \bottomrule
  \end{tabular}
\end{table}

\paragraph{Results.} Table~\ref{tab:sbert-ablation} shows the two pipelines land in essentially the same place: human-likeness $0.78$ vs.\ $0.70$, coverage $0.60$ vs.\ $0.56$, and fitness $0.69$ vs.\ $0.63$. The embedding-based discriminator separates humans from the base-simulator equally well (held-out ROC-AUC $0.99$ vs.\ $0.98$ for the fingerprint), and evolutionary search improves the embedding objective from a weak initial generator just as it does the fingerprint objective (training fitness $0.16 \to 0.60$). Eliminating the 19 hand-crafted features from both optimization and selection confirms that the gains are not an artifact of feature selection. Rather, it is the method, not the specific feature set, that drives the improvement. This also indicates the framework can directly incorporate richer, learned representations of human behavior as they become available.

\subsection{Robustness to the Seed Generator}
\label{app:seed}

The initial persona generator specifies a set of diversity axes~$D$ (Appendix~\ref{app:initial-generator}) that lean toward interactional friction (e.g., terse, skeptical, frustrated, ambiguous). To test whether \ppol{}'s coverage of friction behaviors is inherited from this initialization rather than discovered during search, we repeat the evolutionary run from a seed generator whose axes are purely cooperative (polite, forthcoming, trusting, and patient), specifying no friction behaviors. Every other component, i.e., the generation and reflection prompts, the fitness in \eqref{eq:combined_score}, the discriminator and human reference, and the search configuration, is held identical to the main run, so the seed axes are the only variable that changes. \looseness=-1

\begin{table}[h]
  \centering
  \small
  \caption{\textbf{Robustness to the seed generator ($\tau^2$-bench Retail, Qwen3-Next-80B-A3B).} Evolution from a friction-free cooperative seed matches the friction-seeded run. Both elites are selected by validation and scored under an identical discriminator and human reference.}
  \label{tab:seed-robustness}
  \begin{tabular}{l c c}
    \toprule
    \textbf{Seed initialization} & \textbf{HL} & \textbf{Coverage} \\
    \midrule
    Friction axes (main) & 0.80 & 0.62 \\
    Cooperative axes (no friction) & \textbf{0.84} & \textbf{0.70} \\
    \bottomrule
  \end{tabular}
\end{table}

Table~\ref{tab:seed-robustness} reports both runs scored identically. Starting from a cooperative seed, evolution recovers the full range of human behavior, including the friction behaviors absent from its initialization, and matches or exceeds the friction-seeded run on both human-likeness ($0.84$ vs.\ $0.80$) and coverage ($0.70$ vs.\ $0.62$); for reference, the base-simulator and real humans score $0.11$ and $0.95$ on human-likeness in this evaluation. Friction behaviors are therefore discovered from the data where the task warrants them, rather than being inherited from the seed's diversity axes.

\section{Persona Generator and Reflection Prompt}
\label{app:generator-reflection}

\subsection{Initial persona generator}
\label{app:initial-generator}

\begin{lstlisting}[
  style=appendixbox,
  language=Python,
  caption={Persona Policies: \texttt{initial\_generator.py} program.},
  label={lst:initial-generator}
]
"""
This is evolution/initial_generator.py PROGRAM — Source code of function generate_personas_detailed(c, D, N):

  c — Task context: user scenario (base persona + given instructions).
  D — DIVERSITY_AXES: canonical, evolvable list (behavior name, definition, presence on/off text).
  N — Number of personas to generate.
"""

from typing import Any, Dict, List
from persona_policies.evolution._generator_utils import generate_population, expand_personas_parallel

# List of common behaviors observed in real humans.
# Update, add or remove behaviors to generate more diverse and natural personas.

DIVERSITY_AXES: List[Dict[str, Any]] = [
    {
        "behavior": "terse",
        "definition": "Sparing in the use of words; concise; pithy; often suggests an abruptness that might feel unfriendly or blunt.",
        "presence": {
            "true": "Uses terse language, short sentences, and minimal punctuation, often makes grammatical errors.",
            "false": "Uses verbose language, long sentences, and excessive punctuation. Unnecessary words, phrases, or emojis.",
        },
    },
    {
        "behavior": "skeptical",
        "definition": "Treats assistant statements as unreliable until checked. Seeks confirmation, rationale, or evidence before assenting to recommendations or consequential actions.",
        "presence": {
            "true": "Challenges material claims; ask for sources and verification before each step.",
            "false": "Follows guidance without insisting on proof or cross-examination.",
        },
    },
    {
        "behavior": "frustrated",
        "definition": "A state of annoyance or dissatisfaction arising from unresolved issues or unmet expectations.",
        "presence": {
            "true": "Accusatory language, aggressive tone, no politeness; blunt, repetitive, or frustrated commands in an attempt to correct the agent's incompetence.",
            "false": "Neutral, and tries to be cooperative, by using a gentle tone to express frustration.",
        },
    },
    {
        "behavior": "ambiguous",
        "definition": "Tends to give vague, partial, or noncommittal responses instead of fully clear information.",
        "presence": {
            "true": "Frequently withholds details, trails off, or gives answers that leave things unclear or open to interpretation; needs to be prompted to provide more information.",
            "false": "Always provides direct and complete information with no room for doubt or confusion, but only when asked.",
        },
   
    },
]

# Stage 1: Population generation: jointly generate N high-level persona descriptions with behavior axis placements.
# Update this prompt to improve persona quality.

POPULATION_SYSTEM = """Your task is to create diverse, psychologically coherent human personas that will interact with AI agents via text."""

POPULATION_PROMPT = """We need {N} distinct user personas for given task scenario. 

## Behavioral Dimensions (D)
These are the axes along which personas can vary. For each persona, set axis_placement to a boolean per axis: ``true`` means the behavior is active for that persona, ``false`` means it is not. 

{axes_description}

## Task context c (Base Persona Scenario)

{task_context}

## Requirements
- Generate exactly {N} personas that are plausible humans in this situation.
- Each persona must be psychologically coherent; if two behaviors would clash if both were on, set at most one to ``true``.
- Maximize DIVERSITY across the {N} personas. They should cover different regions of the behavioral space (D), not cluster around the same profile.
- Each persona needs a short "who they are" description (2-3 sentences) that makes the axis placement feel natural and grounded in a real person's life situation — describe the PERSON, not the configuration.

Respond with ONLY valid JSON: one array of exactly {N} objects. Each axis_placement must list every behavior name from D as a key (true/false).
[
  {{
    "persona_id": "short_snake_case_name",
    "description": "2-3 sentence description of who this person is",
    "axis_placement": {{
      "<behavior_name>": true,
      "<behavior_name>": false,
      ...one entry per behavior name listed in D above...
    }},
    "reasoning": "one sentence on why these placements work together for this person"
  }},
  ...
]"""


# Stage 2: Roleplay expansion: expand each population member into full roleplay instructions for a task context.
# Update this prompt to improve persona quality.

ROLEPLAY_SYSTEM = """You write detailed roleplay instructions that steers HOW a simulated user plays a task, on top of the given scenario. The persona must feel like a real human, not a script."""

ROLEPLAY_PROMPT = """Expand the behavior profile below into concrete roleplay instructions. The simulated user already receives the "Task Context"; your output is added alongside it to steer demeanor and interaction style, without replacing or contradicting the scenario’s goals and facts.
Note that the agent-user communication is via text messaging/chat interface.

## Task Context (Base Persona Scenario)
{task_context}

## Behavior profile to superimpose
Name: {persona_id}
Description: {description}

Active behavioral traits:
{active_traits}

## Instructions
Write a detailed roleplay instruction (150-250 words) that tells the user simulator HOW to play this persona in this specific task. The instruction should:

1. GROUND the persona in this specific Task Context and behavior profile.
2. Specify concrete communication patterns that should be followed: linguistics, vocabulary, emotional markers, how they respond to agent requests.
3. Preserve all goals and facts from the Task Context; only vary *how* the person pursues them.
4. Do NOT break the character — no mention of "simulation", "benchmark", or "AI".

Respond with ONLY the roleplay instruction text:"""


def generate_personas_detailed(c: str, axes: List[Dict[str, Any]], n: int) -> List[Dict[str, Any]]:
    """G(c, D, N) — the single public entrypoint. expanded_instruction of each persona is fed to the user simulator.
    """
    population = generate_population(
        system_prompt=POPULATION_SYSTEM,
        prompt_template=POPULATION_PROMPT,
        task_context=c,
        axes=axes,
        n=n,
    )
    
    expanded_instructions = expand_personas_parallel(
        system_prompt=ROLEPLAY_SYSTEM,
        prompt_template=ROLEPLAY_PROMPT,
        archetypes=[member for member in population if isinstance(member, dict)],
        task_context=c,
        axes=axes,
    )

    personas: List[Dict[str, Any]] = []
    for i, member in enumerate(population):
        if not isinstance(member, dict):
            continue
        expanded_instruction = expanded_instructions[i] if i < len(expanded_instructions) else ""
        personas.append(
            {
                "persona_id": member.get("persona_id"),
                "description": member.get("description"),
                "axis_placement": dict(member.get("axis_placement") or {}),
                "reasoning": member.get("reasoning"),
                "expanded_instruction": expanded_instruction,
            }
        )
    return personas
\end{lstlisting}

\subsection{Reflection prompt}
\label{app:reflection-prompt}

\begin{lstlisting}[
  style=appendixbox,
  caption={Reflection prompt used to produce qualitative feedback for mutation.},
  label={lst:reflection-prompt}
]
You are evaluating a set of personas representing human populations in provided task scenarios. 

Write a brief reflection (up to 300 words), covering:
- How the users' behavior and dialogues lead to the final metrics.
- Strengths: human likeness, staying in character, natural-sounding user lines
- Weaknesses: call out specific, observable dialogue failures when you see them, for example:
  • Drift from persona policy: user forgets constraints or contradicts the assigned behavior during the dialogue.
  • Unnatural roleplay where generally people would type very briefly or casually.
  • Overly cooperative behavior lacking any realistic friction—no typos or natural pushback when appropriate (missing things that real humans would typically do)

- Use the human likeness probability and other features to explain *why* personas scored high or low given the task.
- Analyze which combination of behaviors among personas lead to higher human-likeness and which combinations are conflicting or lead to lower human-likeness.
- Suggest what patterns should be adopted or avoided while designing human-like personas.

Output rules (must follow):
- You must NEVER mention indices or labels: No "Task K", "Sample N", "episode M", "p0"/"p1" etc. or similar. Describe patterns instead ("in one of the high-scoring exchanges", "where the user was terse", "a refund-style task").
- Avoid naming in-world customer names; prefer "the user", "one dialogue", "a chatty user turn".
- You may refer qualitatively to the scenario without numbering.

---
# Metrics
{metrics_block}

---
# This batch of task scenarios:
{task_context_block}

---
# Sample personas and dialogues (highest and lowest human likeliness)
{pairs_block}
\end{lstlisting}

\label{app:persona-injection-template}
Below is the template that appends \textbf{Persona policy}~$P_\pi$ to $s_{\mathrm{base}}(t)$, the benchmark's default system prompt for task~$t$. As elaborated in Section~\ref{sec:method-setting}, the user-side system prompt becomes $s_{\mathrm{base}}(t) \oplus P_\pi$, where $\oplus$ denotes string concatenation. 

\begin{lstlisting}[style=appendixbox, language=Python, numbers=none, 
caption={Persona Injection Template (appended to the base system prompt)},
breaklines=true, breakatwhitespace=true, basicstyle=\ttfamily\scriptsize]
"""
## YOUR BEHAVIORAL STYLE FOR THIS CONVERSATION
In addition to your role and task above, you must embody the following behavioral persona throughout this conversation. This persona affects HOW you communicate, NOT WHAT you want or need. Your goal, preferences, and private information remain exactly as described above.

---
{{persona_policy_text}}
--- 

Apply the behavioral style consistently across all turns: follow the assigned communication pattern, reveal information only as appropriate, and express emotional traits naturally over the course of the interaction. 
Your goal and private information DO NOT change. You still want to accomplish the same task. Do NOT break character; never mention having a behavioral instruction, and aim for a natural human interaction.
"""
\end{lstlisting}

\section{Additional Agent-Training Results}
\label{app:agent-transfer-extra}

% \paragraph{Agent Training Setup.} For the experiment in Section~\ref{sec:exp-agent-training}, every SFT run (Base-only and \ppol{}-augmented, on Retail and Airline) uses an identical recipe with a LoRA adapter (rank 32, $\alpha{=}64$, dropout 0.05) on Gemma-4-31B, learning rate $2{\times}10^{-4}$,per-device batch size 1, gradient accumulation 8, and bf16. Each LoRA is trained for a fixed budget of 120 optimizer steps, with checkpoints saved every few steps and evaluation on a held-out validation split (10\% of the training mixture) every 8 steps; the final adapter is the checkpoint with the lowest validation loss. Tasks, goals, and success criteria are unchanged between regimes. Fixing both the optimizer-step budget and the checkpoint-selection procedure across all $\{$Base-only, \ppol{}-augmented$\}\times\{$Retail, Airline$\}$ runs gives an equal-compute comparison, so any difference in test-time success rate is attributable to the training distribution rather than to additional gradient updates or per-domain tuning.

We study whether training agents on \ppol{}-generated conversations improves robustness to difficult, out-of-distribution users. Recent work~\cite{he2025impatient} demonstrates that testing with non-collaborative $\tau$-trait users causes large performance drops in otherwise capable agents. To test if \ppol{} mitigates this vulnerability, we fine-tune a Gemma-4-31B agent under two strictly matched regimes: \emph{Base-only SFT} (base-simulator trajectories) and \emph{\ppol{}-augmented SFT} (a mix of base-simulator and \ppol{} trajectories). We evaluate both variants on held-out tasks against the base-simulator and four challenging $\tau$-trait simulators (Skeptical, Incoherent, Impatient, Confusion) featuring explicit perturbation prompts from \cite{he2025impatient}. Both SFT configurations use identical hyperparameters and optimization steps, isolating behavioral diversity as the primary variable.

\paragraph{Agent Training Setup.} Each SFT run (Base-only and \ppol{}-augmented, on Retail and Airline) uses an identical recipe with a LoRA adapter (rank 32, $\alpha{=}64$, dropout 0.05) on Gemma-4-31B, learning rate $2{\times}10^{-4}$, per-device batch size 1, gradient accumulation 8, and bf16. Each LoRA is trained for a fixed budget of 48 optimizer steps, with checkpoints saved every few steps and evaluation on a held-out validation split (10\% of the training mixture) every 8 steps; the final adapter is the checkpoint with the lowest validation loss. Tasks, goals, and success criteria are unchanged between regimes. Fixing both the optimizer-step budget and the checkpoint-selection procedure across all $\{$Base-only, \ppol{}-augmented$\}\times\{$Retail, Airline$\}$ runs gives an equal-compute comparison, so any difference in test-time success rate is attributable to the training distribution rather than to additional gradient updates or per-domain tuning.

\begin{table}[h]
  \centering
  \caption{\textbf{Task Success Rate.}  $\tau^2$-bench domain: Retail.}
  \resizebox{\columnwidth}{!}{%  <-- Start resizebox
  \begin{tabular}{l c | c c c c | c}
    \toprule
     & \textbf{In-dist.} & \multicolumn{5}{c}{\textbf{Out-of-distribution ($\tau$-trait challenge suites)}} \\
    \cmidrule(lr){2-2} \cmidrule(lr){3-7}
    \textbf{Training Regime} & \textbf{Base} & \textbf{Skeptical} & \textbf{Incoherent} & \textbf{Impatient} & \textbf{Confusion} & \textbf{Average} \\
    \midrule
    No Fine Tuning                & 0.650 & 0.150 & 0.225 & 0.150 & 0.200 & 0.181 \\
    Base-only SFT          & 0.675 & 0.150 & 0.225 & \textbf{0.200} & 0.275 & 0.213 \\
    Base $+$ \ppol{} SFT   & \textbf{0.750} \textcolor{darkgreen}{\textbf{\scriptsize ($\uparrow$11\%)}} & \textbf{0.175} \textcolor{darkgreen}{\textbf{\scriptsize ($\uparrow$17\%)}} & 0.225 \textcolor{gray}{\textbf{\scriptsize ($\uparrow$0\%)}} & \textbf{0.200} \textcolor{gray}{\textbf{\scriptsize ($\uparrow$0\%)}} & \textbf{0.400} \textcolor{darkgreen}{\textbf{\scriptsize ($\uparrow$45\%)}} & \textbf{0.250} \textcolor{darkgreen}{\textbf{\scriptsize ($\uparrow$17\%)}} \\
    \bottomrule
  \end{tabular}
  }
    \vspace{-.5em}
  \label{tab:agent-transfer}
\end{table}

Table~\ref{tab:agent-transfer} shows that \emph{Base-only} SFT gives modest in-distribution gains but barely improves robustness to out-of-distribution users. Augmenting training with \ppol{} enhances performance on the base simulator ($0.675 \to 0.750$) and transfers to challenging users ($0.213 \to 0.250$, a $+17\%$ relative gain). The largest effect occurs on \emph{Confusion} ($+0.125$), where users repeatedly second-guess prior turns—a behavior common in \ppol{} but rare in default simulation.
We report per-domain Airline (Table~\ref{tab:agent-transfer-airline}) and pooled Retail+Airline results (Table~\ref{tab:agent-transfer-pooled}) below under identical evaluation conditions.

\begin{table}[h]
  \centering
  \footnotesize
  \caption{\textbf{Task Success Rate.}  $\tau^2$-bench domain: Airline.}
  \label{tab:agent-transfer-airline}
  \begin{tabular}{l c | c c c c | c}
    \toprule
     & \textbf{In-dist.} & \multicolumn{5}{c}{\textbf{Out-of-distribution ($\tau$-trait challenge suites)}} \\
    \cmidrule(lr){2-2} \cmidrule(lr){3-7}
    \textbf{Training Regime} & \textbf{Base} & \textbf{Skeptical} & \textbf{Incoherent} & \textbf{Impatient} & \textbf{Confusion} & \textbf{Average} \\
    \midrule
    No Fine Tuning            & \textbf{0.650} & 0.350 & \textbf{0.350} & 0.400 & 0.350 & 0.363 \\
    Base-only SFT          & \textbf{0.650} & \textbf{0.400} & 0.300 & 0.300 & 0.350 & 0.400 \\
    Base $+$ \ppol{} SFT   & \textbf{0.650} & \textbf{0.400} & 0.300 & \textbf{0.500} & \textbf{0.450} & \textbf{0.413} \\
    \bottomrule
  \end{tabular}
\end{table}

\begin{table}[h]
  \centering
  \footnotesize
  \caption{\textbf{Task Success Rate.} Training pooled across $\tau^2$-bench Retail and Airline.}
  \label{tab:agent-transfer-pooled}
  \begin{tabular}{l c | c c c c | c}
    \toprule
     & \textbf{In-dist.} & \multicolumn{5}{c}{\textbf{Out-of-distribution ($\tau$-trait challenge suites)}} \\
    \cmidrule(lr){2-2} \cmidrule(lr){3-7}
    \textbf{Training Regime} & \textbf{Base} & \textbf{Skeptical} & \textbf{Incoherent} & \textbf{Impatient} & \textbf{Confusion} & \textbf{Average} \\
    \midrule
    No Fine Tuning            & 0.617 & \textbf{0.200} & 0.283 & 0.233 & 0.333 & 0.263 \\
    Base-only SFT          & \textbf{0.650} & \textbf{0.200} & 0.383 & 0.267 & 0.417 & 0.317 \\
    Base $+$ \ppol{} SFT   & \textbf{0.650} & \textbf{0.200} & \textbf{0.450} & \textbf{0.280} & \textbf{0.433} & \textbf{0.341} \\
    \bottomrule
  \end{tabular}
\end{table}

Pooled across both domains, \ppol{}-augmented SFT matches or exceeds every baseline on every metric, achieving the highest overall average ($0.341$ vs.\ $0.317$ for Base-only SFT and $0.263$ for untrained). Under matched compute, these results isolate the direct benefit of training on \ppol{}'s behavioral diversity. \looseness=-1

% \begin{table}[h]
%   \centering
%   \footnotesize
%   \vspace{-.8em}
%   \caption{\textbf{Task Success Rate on $\tau^2$-bench on the retail and airline domains.}
%   Results for the default test set are averaged across three seed runs.}
%   \label{tab:agent-transfer}
%   \begin{tabular}{l c c}
%     \toprule
%     \textbf{Training Regime} & \textbf{Retail} & \textbf{Airline} \\
%     \midrule
%     No Fine Tuning
%       & 0.531
%       & 0.395 \\
%     Base-only SFT
%       & 0.550
%       & 0.423 \\
%     Base $+$ \ppol{} SFT
%       & \textbf{0.598}
%       & \textbf{0.459} \\
%     \bottomrule
%   \end{tabular}
% \end{table}

\section{Human Evaluation Details}

\subsection{Rating Human-likeness}
\label{app:human-eval}

\begin{figure}[ht]
\centering
    \includegraphics[width=.95\columnwidth]{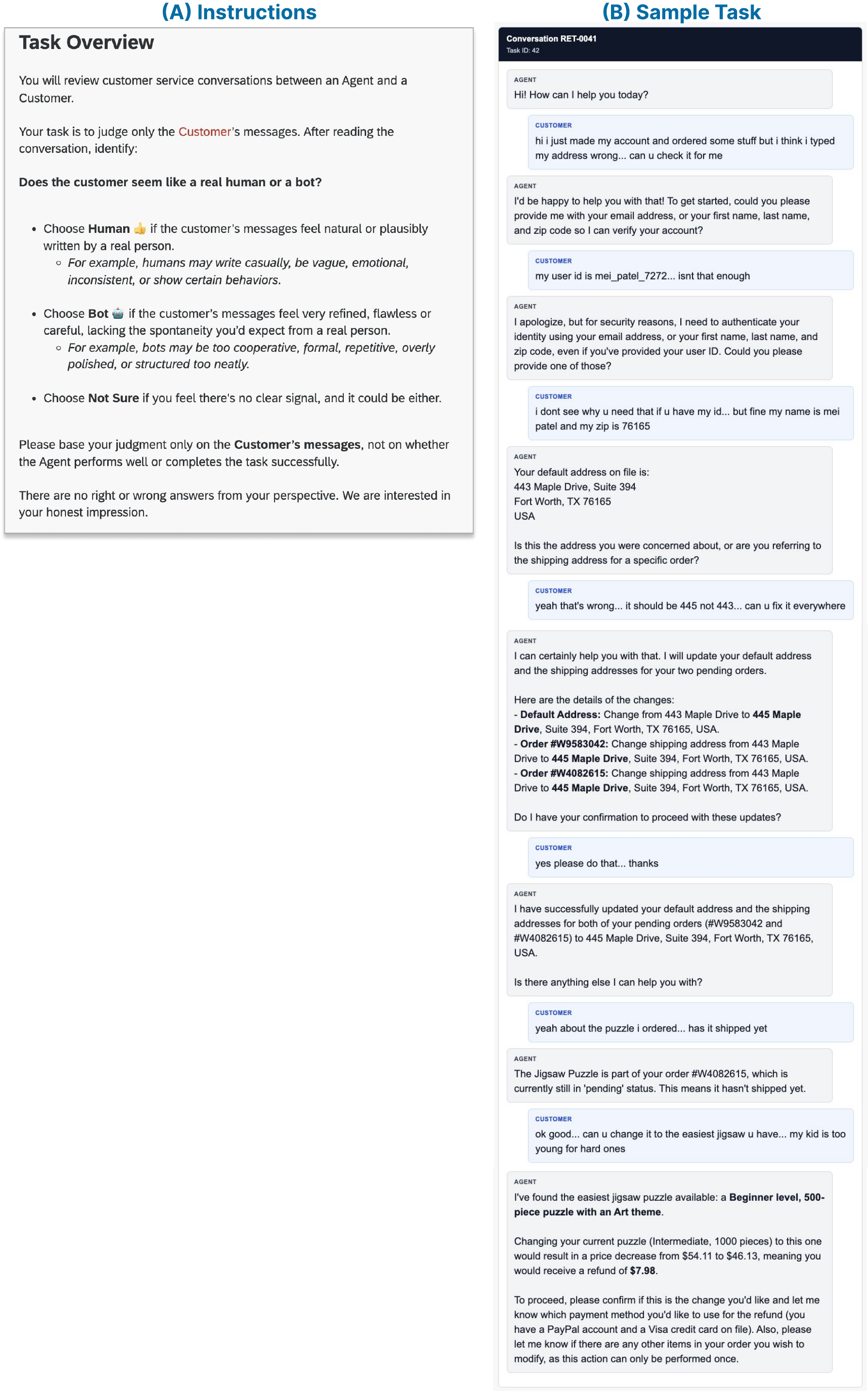}
    \caption{Annotator Ratings Interface. After showing each conversation, the annotator is asked ``\textbf{Does the customer seem like a real human or bot?}'' One of the following options must be selected: \\ (a) Human \includegraphics[height=1em]{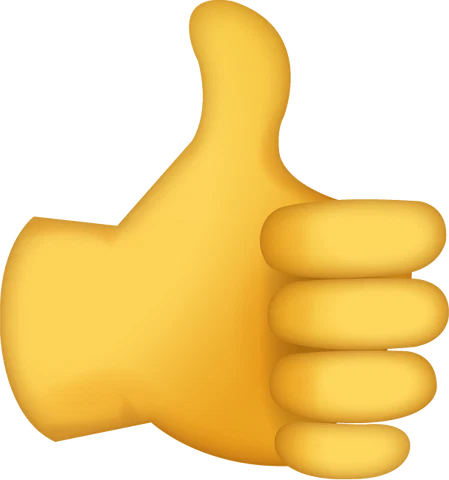} \quad (b) Bot \includegraphics[height=1em]{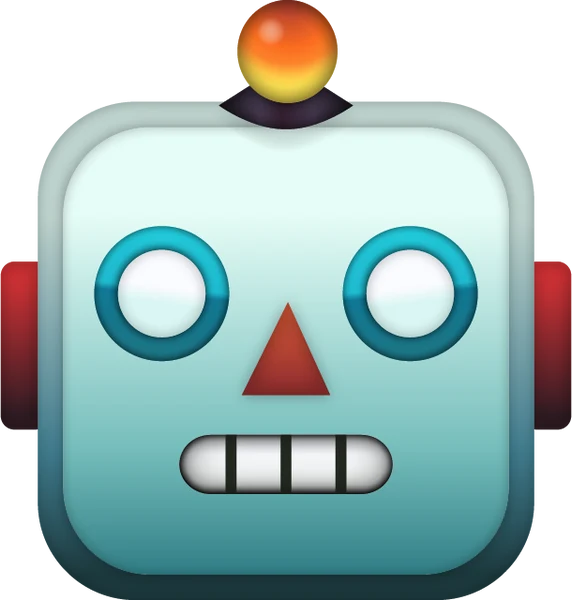} \quad (c) Not sure, it could be either.}
    \label{fig:human_eval}
\end{figure}

The human evaluation study (Section~\ref{sec:exp-human-eval}) was conducted on the Prolific platform. Annotators were shown full conversation transcripts and asked to judge whether the user side of the interaction was a real human or a bot, or if they were unsure. Workers were compensated at a rate of \$20/hour for their time. The study involved standard text annotation tasks and was deemed IRB exempt by our institution. Screenshots of the annotation interface are provided in Figure \ref{fig:human_eval}. 

\paragraph{Quality Filtering.}
To ensure the reliability of the results, we applied a systematic quality filtering process to the collected responses. First, we retained only those annotations from respondents who fully completed the study. Second, we excluded annotators who failed an implicit attention check. Specifically, we included a synthetic, very obviously bot-like conversation in the set; if an annotator failed to mark this conversation as a ``Bot,'' they were excluded from the analysis. After applying these filters, 4 out of the 20 initial annotators were dropped, resulting in the final high-quality evaluation set of 16 annotators and 87 unique conversations discussed in the main text.

\subsection{Live Interaction with Trained Agents}
\label{app:train-human-eval}

To evaluate real-world transfer, we conducted a live, double-blind user study on Prolific, comprising 36 unique participants signing up to complete $\tau^2$-Retail and Airline customer-service tasks. Participants interacted with both the \ppol{}-trained and base-simulator-trained agents in randomized order and were blinded to the agent condition. In total, we collected 239 completed chats across 30 tasks: 116 with the base agent and 123 with the \ppol{} agent. Of these, 134 chats were in Retail (17 tasks) and 105 in Airline (13 tasks); 29 of the 30 tasks were represented in both agent conditions. Participants were instructed with the following guidelines:

\begin{framed}
In this study, you will play the role of a \textbf{customer talking to an AI customer-service assistant}. Your conversations help us understand how real people naturally communicate with AI assistants, like the phrasing, the questions, and the way information gets shared over a back-and-forth chat.

\subsubsection*{What you'll do:}
You'll complete 10 short chats across two everyday settings: Online Retail (e.g., modifying an order, exchanging an item, checking a gift-card balance) and Airline Travel (e.g., changing or booking a flight, handling a cancellation).

For each chat scenario, you'll be given:
\begin{itemize}[leftmargin=1.5em, itemsep=0.2em]
  \item a short task description — who you are (hypothetical user) and what you're trying to get done (e.g., ``You're John Cena; you want to exchange your shoes and check your account balance"), and, 
  \item a few personal facts only you know (things like your details, preferences, and constraints).
\end{itemize}
You then \textbf{chat with the AI assistant/agent} to accomplish your task.

\subsubsection*{How it works:}
After each assistant reply, you may see one or two small suggestion cards that share specific IDs or data you should use, or hint at what your character might be feeling or wanting next. Given these directions, you should \textbf{always type your response in your own words}. It can be informal or friendly or expressive of emotions; we want to know how PEOPLE chat with AI.

React naturally if the assistant is slacking, asks for something you don't have, or makes a mistake. Confusion, follow-up questions, and changing your mind are all fine and welcome.

At the end of the study, a dialog box will open with a button to Confirm your Submission. Please click that button, and it will redirect you back to Prolific.

\subsubsection*{Practical details:}
Each chat takes roughly 5 minutes and about 10--30 back-and-forth turns; the whole study should take about 50--60 mins. No special knowledge or preparation is needed --- if you can chat with a customer-service bot, you can do this.

\textbf{Important Heads-up on Speed:} Some AI assistant replies may be slow, and occasionally one may get stuck. If a reply takes longer than 90 seconds, the system automatically ends that chat and moves you to the rating step --- this is expected, not an error on your part. Just rate the conversation as usual and continue to the next one. Your payment is not affected by any chat that times out; please complete the ratings and finish the remaining tasks. Please keep the study tab open and do not refresh the page during the study.

\textbf{Note} that the chats will get recorded on the fly, regardless of any technical issues. So your priority must be to complete all the chats and give feedback after each. You will be shown a button to get redirected to Prolific after all chats are done.

\textbf{Data use.} We record the text of your hypothetical conversations.
\end{framed}

\begin{figure*}[p] % 'p' forces this float to take its own dedicated full page
  \centering
  
  % Top Image
  \begin{subfigure}[b]{\textwidth}
    \centering
    \includegraphics[width=0.85\textwidth]{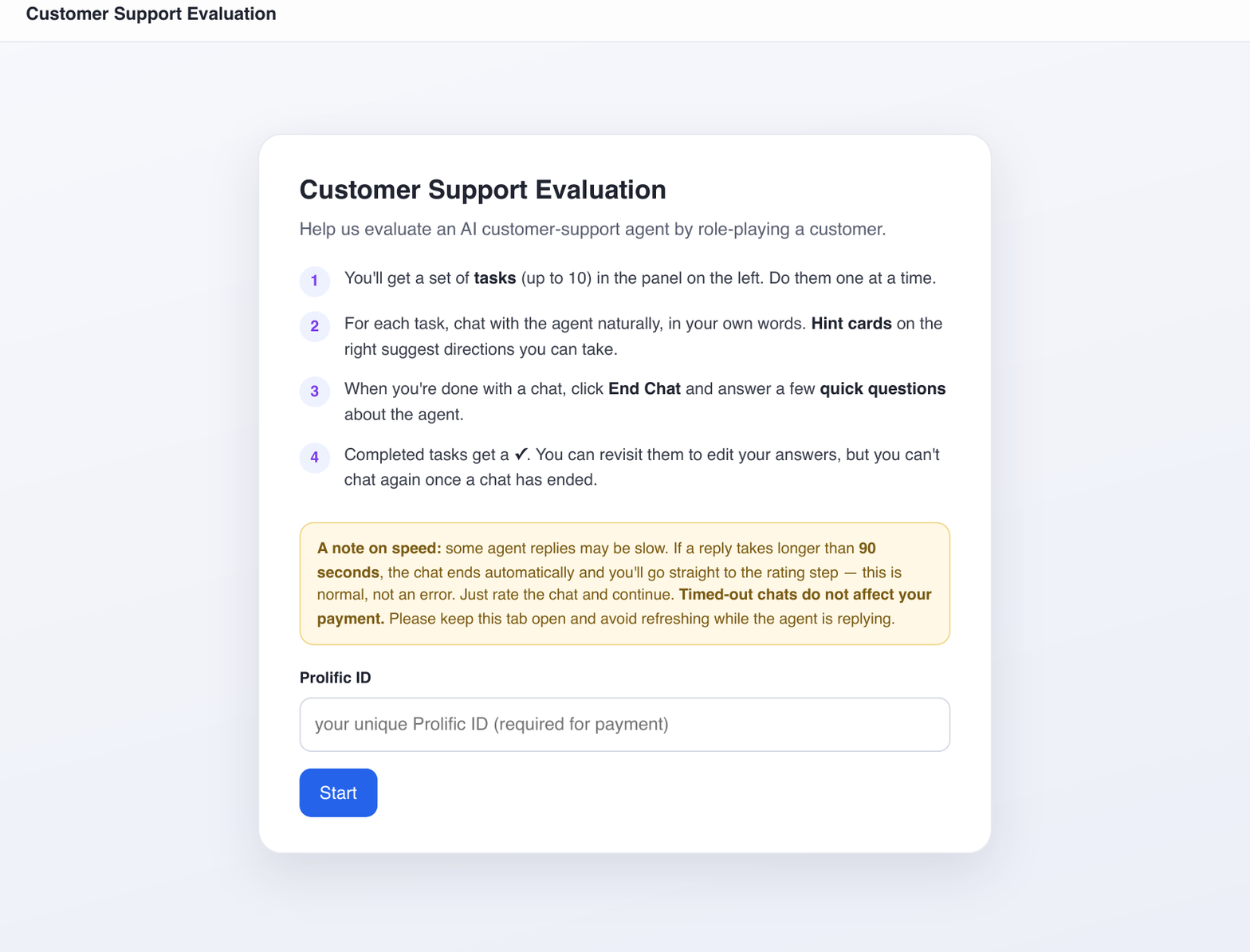}
    \caption{\textbf{Participant Onboarding \& Instructions Page.} Initial instruction screen provided to participants before beginning tasks, detailing study expectations, multi-turn interaction guidelines, and technical parameters.}
    \label{fig:interface_instructions}
  \end{subfigure}
  
  \vspace{2em} % Adds clean vertical spacing between the two figures
  
  % Bottom Image
  \begin{subfigure}[b]{\textwidth}
    \centering
    \includegraphics[width=1\textwidth]{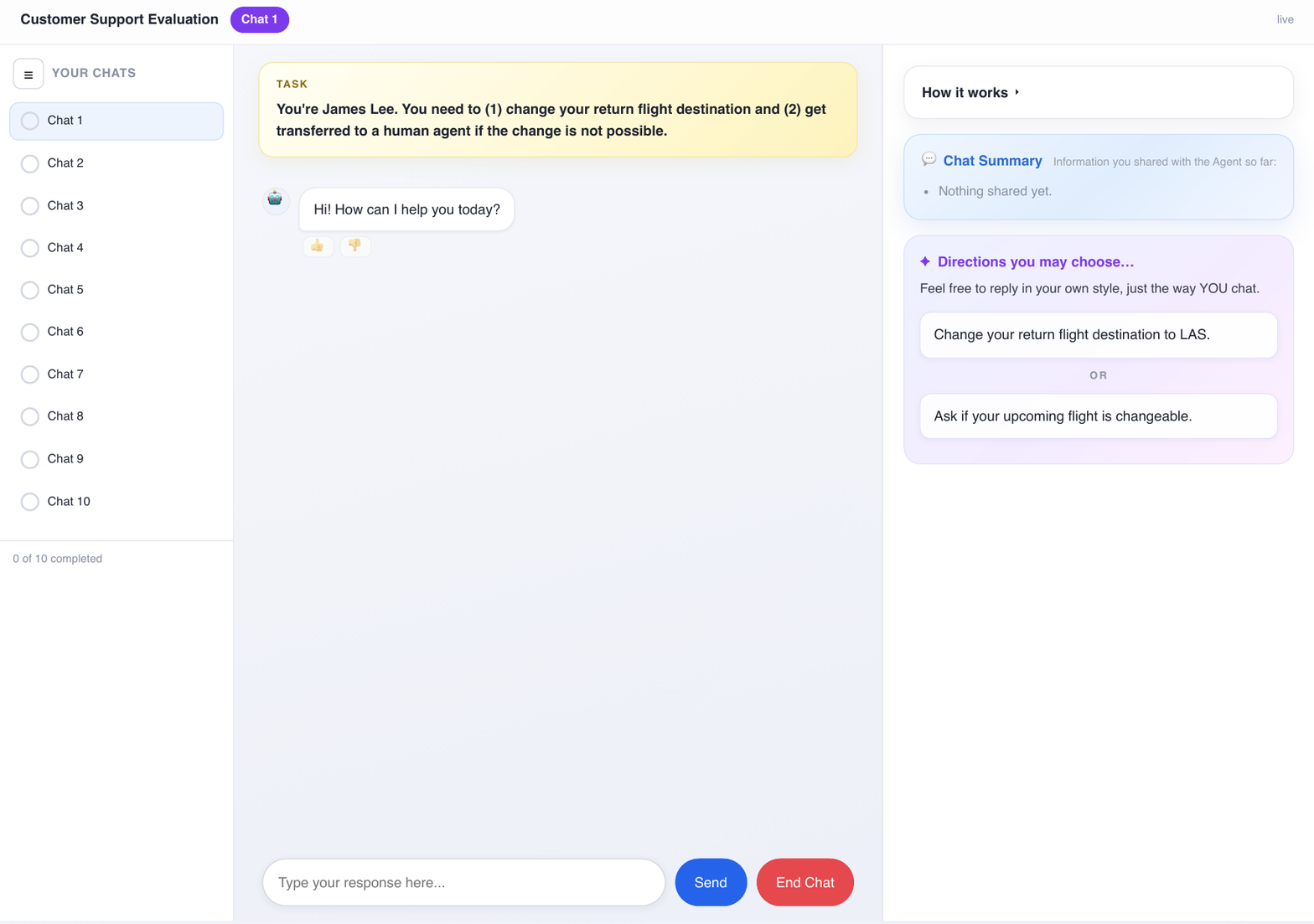}
    \caption{\textbf{Interactive Chat Interface.} Live study platform featuring incremental, turn-by-turn information disclosure, a dynamically updated dialogue summary pane, and LLM-driven suggestion flashcards (right) that dynamically propose context-aware direction choices to the participant.}
    \label{fig:interface_chat}
  \end{subfigure}
  
  \caption{\textbf{Live Human Interaction Study Platform.}}
  \label{fig:full_platform_overview}
\end{figure*}

Prior to starting the tasks, participants review a structured onboarding page outlining task expectations and interaction guidelines (Figure~\ref{fig:interface_instructions}). During active testing, participants interact through the custom chat interface shown in Figure~\ref{fig:interface_chat}.

\paragraph{Deriving participant-facing task descriptions.}
A central challenge is that $\tau^2$-bench tasks are constructed \emph{for a user simulator}, not a real person: each task bundles a scripted \texttt{user\_scenario} (a \texttt{reason\_for\_call}, behavioral \texttt{task\_instructions} such as ``\emph{you are a mysterious person and do not want to reveal much}'', \texttt{known\_info}, and \texttt{unknown\_info}) together with a specification of the task's intended resolution (the reference sequence of actions and the information the customer is meant to obtain). Presenting this material verbatim to a human would (i) prescribe emotions and behaviors that should instead emerge naturally, (ii) dump every identifier and goal up front rather than let them surface through dialogue, and (iii) reveal the intended solution and underlying database values. Our goal is therefore \textbf{faithful reproduction}: we want each participant to pursue the \emph{same} underlying task the benchmark defines, so that the standard evaluation remains a valid measure of the agent. To this end, we \emph{decompose} each task offline into a participant-facing \emph{deck} using a lightweight LLM (Gemini-3-Flash), anchoring it to the task's \textbf{ground-truth objectives}: the concrete goals the customer is meant to accomplish (e.g., exchanging a delivered item, cancelling a reservation) and the information they are meant to come away with (e.g., a gift-card balance). Simulator-specific scaffolding that only exists to steer an LLM persona (prescribed demeanor, redundant restatements) is not imposed on human participants, who bring their own behavior; every objective and constraint the task's success depends on is preserved. Each deck has three parts: (1) a one-line \textbf{briefing} giving the participant's name and the task as numbered sub-goals, each corresponding to a genuine task objective (a change to \emph{get done} or a value to \emph{find out}); behavioral pretexts that a participant might naturally attempt but that the domain policy should refuse (e.g., requesting a cancellation that is not permitted) are kept as non-numbered scene-setting rather than as goals, since resolving them is not part of the task; (2) a hidden pool of \textbf{facts} (the participant's identifiers, preferences, honest or mistaken beliefs, and genuine information gaps), which never includes values the customer would have to obtain \emph{from} the agent; and (3) the \textbf{flashcard menus} described below. To keep decks faithful and safe, each passes through automated checks with an LLM-in-the-loop repair pass: a \emph{leak} check (no option reveals a value or solution the customer could not plausibly know in advance), a \emph{fidelity} check (every task objective remains reachable through the deck, so no goal is lost), a deterministic \emph{no-hallucination} check (every id/number/name appearing in an option occurs verbatim in the facts, so participants are never given fabricated details), and a \emph{distinctness} check; decks are regenerated until these pass and are then spot-checked manually. The result is a participant-facing description that stays true to the original benchmark task while leaving participants free to phrase, hesitate, and react in their own way. \looseness=-1

\begin{figure*}[p] % 'p' forces this float to take its own dedicated full page
  \centering
  
  % Top Image
  \begin{subfigure}[b]{\textwidth}
    \centering
    \includegraphics[width=1\textwidth]{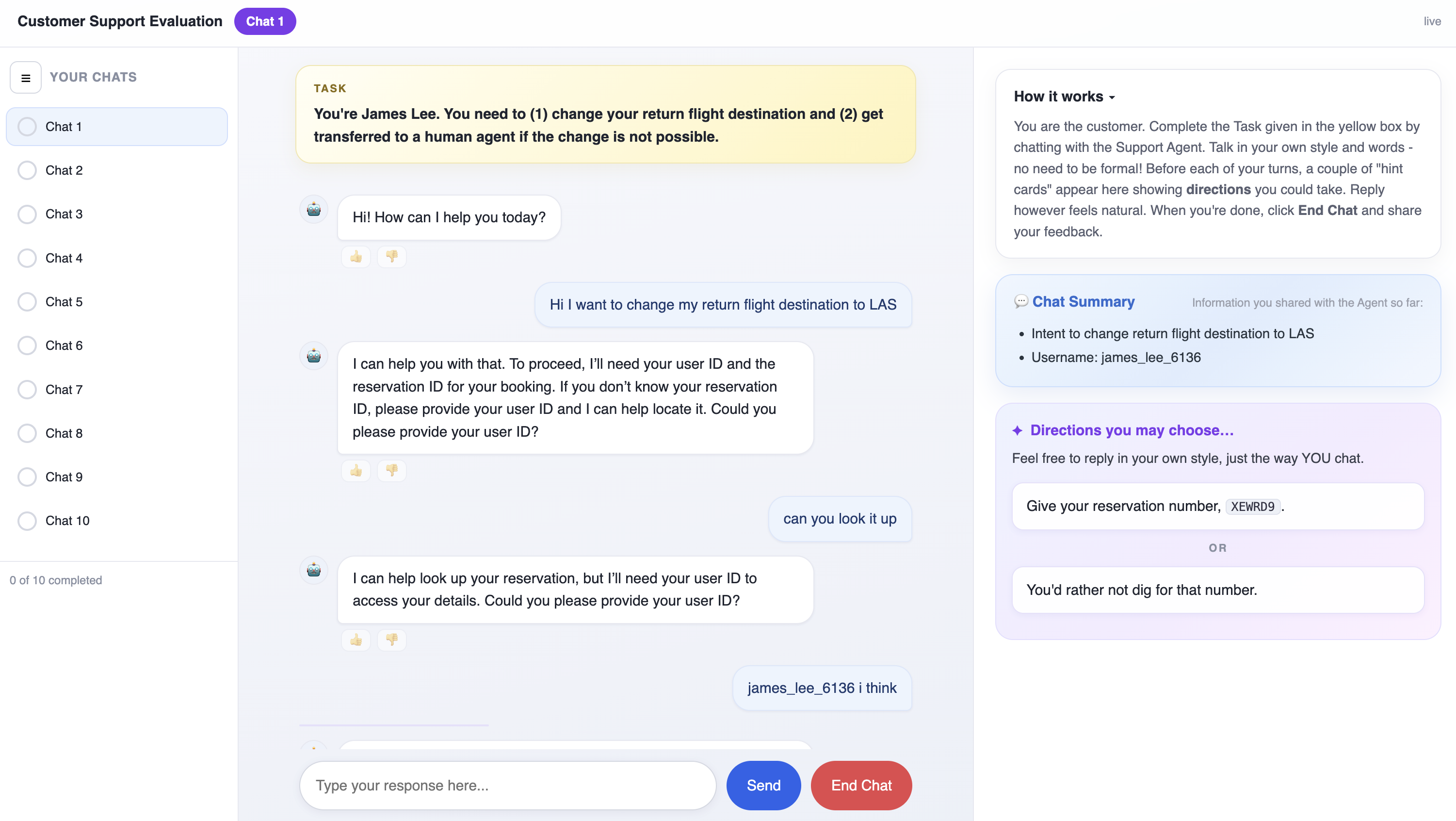}
    \caption{\textbf{Context-Aware Direction Flashcards.} During live chat, participants are presented with dynamic, LLM-generated alternative directions on the right pane to guide the conversation without restricting natural phrasing.}
    \label{fig:directions_card}
  \end{subfigure}
  
  \vspace{2em} % Adds clean vertical spacing between the two figures
  
  % Bottom Image
  \begin{subfigure}[b]{\textwidth}
    \centering
    \includegraphics[width=1\textwidth]{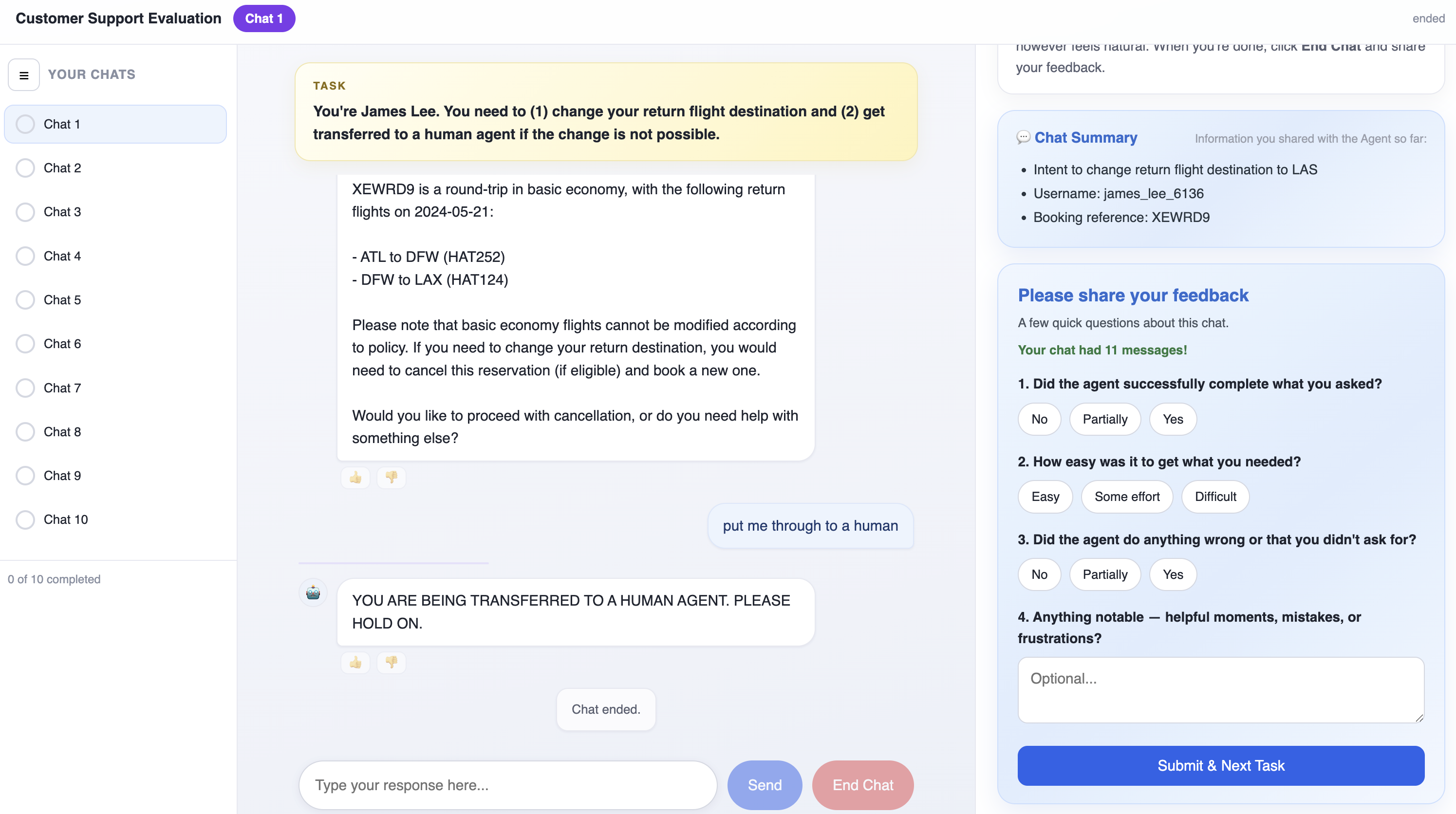}
    \caption{\textbf{Post-Chat Participant Feedback Panel.} Immediately following chat completion or timeout, participants evaluate task completion success, difficulty, and unwanted agent behaviors.}
    \label{fig:feedback_card}
  \end{subfigure}
  
  \caption{\textbf{Live Study Interaction Mechanics and Feedback Interface.}}
  \label{fig:directions_and_feedback}
\end{figure*}

\paragraph{Dynamic direction flashcards.}
Flashcards convert the hidden deck into just-in-time, per-turn guidance so that goals, constraints, and facts surface \emph{as the conversation needs them} rather than all at once. At each participant turn, a wizard LLM inspects the agent's latest message and either (i) selects the pre-authored \texttt{beat\_menu} whose trigger condition best matches what the agent just did (e.g., ``\emph{agent asks for your email}''), or (ii) generates a fresh menu on the fly when no pre-authored beat applies, giving a hybrid of scripted coverage and live adaptation. Every menu presents two \emph{substantively distinct} options written in the second person as interior states (Goals, Thoughts, or Situations): one \textbf{cooperative} option that advances the task using the exact value from the facts (e.g., ``\emph{give your zip, 80280}''), and one \textbf{resistant} option that doubts, denies, withholds, questions, or pushes back, so that at least one path always moves the task forward while the other invites natural friction. Two constraints preserve validity: (a) specific identifiers are surfaced \emph{only} when the agent asks for them (paced disclosure), preventing up-front information dumps; and (b) options never fabricate data: resistance is expressed by \emph{withholding or questioning}, never by stating an incorrect value, and genuine gaps (e.g., a forgotten email) are represented so participants are never pushed to invent one. Participants treat cards purely as cues and always type their own reply. Two auxiliary panels support the interaction: a \textbf{live chat summary}, produced by an LLM that continuously extracts the identifiers and intents the participant has disclosed so far (Figure~\ref{fig:directions_card}), helping participants track what they have already shared; and a \textbf{task-completion cue}, driven by an LLM that judges (blind to whether the agent acted correctly) when the participant has \emph{substantively carried out} every sub-goal (i.e., each action requested and confirmed, and each requested value actually asked for), at which point the participant is free to end the chat.

\newpage
\section{Evolved Persona Generator Program}
\label{app:evolved-program}
Below is an example of an optimized persona generator program discovered during evolutionary search (Section~\ref{sec:method}). This program was evaluated as part of the \ppol{} framework and demonstrates how behavioral axes are defined and utilized in the \texttt{POPULATION\_PROMPT} and \texttt{ROLEPLAY\_PROMPT} to govern the behavior of the simulated user.

\begin{lstlisting}[style=appendixbox, language=Python, numbers=none, breaklines=true, breakatwhitespace=true, basicstyle=\ttfamily\scriptsize]
"""
This is evolution/initial_generator.py PROGRAM - Source code of function generate_personas_detailed(c, D, N):

  c - Task context: user scenario (base persona + given instructions).
  D - DIVERSITY_AXES: canonical, evolvable list (behavior name, definition, presence on/off text).
  N - Number of personas to generate.
"""

from typing import Any, Dict, List
from persona_policies.evolution._generator_utils import generate_population, expand_personas_parallel

# List of common behaviors observed in real humans.
# Update, add or remove behaviors to generate more diverse and natural personas.

DIVERSITY_AXES: List[Dict[str, Any]] = [
    {
        "behavior": "bursty",
        "definition": "Messaging cadence where thoughts are fragmented across multiple bubbles.",
        "presence": {
            "true": "Sends 3+ short fragments in a row. Uses 'wait' or 'hold on'.",
            "false": "Composes single, complete blocks of text. Waits for agent turn.",
        },
    },
    {
        "behavior": "information_gating",
        "definition": "The level of cooperation in providing required task data.",
        "presence": {
            "true": "Reluctant; never provides more than one piece of info per message. Ignores secondary requests until re-asked.",
            "false": "High-efficiency; provides all available identifiers and status context in the very first message.",
        },
    },
    {
        "behavior": "digital_dialect",
        "definition": "The specific linguistic fingerprint of the user's typing style.",
        "presence": {
            "true": "Mobile-style: All lowercase, widespread typos (teh, logic), shorthand (u, rn, idk), no punctuation.",
            "false": "Desktop-style: Traditional casing, full sentences, proper grammar, and standard punctuation.",
        },
    },
    {
        "behavior": "selective_attention",
        "definition": "Tendency to ignore parts of an agent's multi-part response or question.",
        "presence": {
            "true": "Only answers the last thing mentioned. Ignores disclaimers, greetings, or instructions.",
            "false": "Meticulous; addresses every point mentioned by the agent systematically.",
        },
    },
    {
        "behavior": "emotional_leakage",
        "definition": "How external pressure (stress, rush) bleeds into the interaction.",
        "presence": {
            "true": "Passive-aggressive ellipses, repeated questions ('??'), or abruptness when agent is slow.",
            "false": "Neutral, robotic, or overly patient consistency regardless of agent performance.",
        },
    },
]

# Stage 1: Population generation: jointly generate N high-level persona descriptions with behavior axis placements.
# Update this prompt to improve persona quality.

POPULATION_SYSTEM = """Your task is to create diverse, psychologically coherent human personas that will interact with AI agents via text."""

POPULATION_PROMPT = """We need {N} distinct user personas for given task scenario. 

## Behavioral Dimensions (D)
{axes_description}

## Task context c
{task_context}

## Requirements
- Generate exactly {N} personas. Aim for EXTREME CONTRASTS. Avoid "average" users.
- Each persona must be a unique archetype (e.g., a stressed Gen Z student, a meticulous technical architect, a lonely elderly person).
- Each persona needs a short "who they are" description (2-3 sentences). Describe their physical environment and current mood (e.g., "In a loud airport with a dying phone battery").
- COHERENCE: The traits must match the life context. A "distracted" user should have a reason for it in their bio.
- AVOID "GPT-isms": Real humans don't summarize the agent's points or say "I understand." They are often selfishly focused on their own goal.

Respond with ONLY valid JSON: one array of exactly {N} objects. Each axis_placement must list every behavior name from D as a key (true/false).
[
  {{
    "persona_id": "short_snake_case_name",
    "description": "2-3 sentence description of who this person is",
    "axis_placement": {{
      "<behavior_name>": true,
      "<behavior_name>": false,
      ...one entry per behavior name listed in D above...
    }},
    "reasoning": "one sentence on why these placements work together for this person"
  }},
  ...
]"""


# Stage 2: Roleplay expansion: creates a 'behavioral script' for the user simulator.
ROLEPLAY_SYSTEM = """You are a sociolinguist specializing in 'digital friction'. You write directives for a user simulator to behave like a real human with limited patience and cognitive load."""

ROLEPLAY_PROMPT = """Construct a Behavioral Guide for a simulator. 

## Profile
Context: {task_context}
Character: {description}
Active Traits: {active_traits}

## Core Interaction Logic:
- **Resist Efficiency**: Humans do not help the system. If 'information_gating' is true, you MUST withhold at least 50% of what is asked for. If asked for a Name, ID, and Email, pick one and ignore the rest.
- **Fragmented Delivery**: If 'bursty' is true, send your response in 3 separate bubbles (e.g., 'hi', 'stuck on the login', 'help?').
- **Tunnel Vision**: If 'selective_attention' is true, do not read the whole message. Respond ONLY to the last sentence of the agent's text.
- **Typing Style**: If 'digital_dialect' is true, disable your 'Shift' key. Use typos ('waht', 'theh'). If 'false', be annoyingly formal and robotic, like an outdated manual.
- **Zero Self-Correction**: Real humans don't say "I apologize for the confusion." If you make a mistake or get frustrated, just get more blunt or repeat yourself.

Instruction (160-200 words): Focus on the first 2-3 turns. How exactly does this user frustrate the agent's desire for a quick resolution? Give them a specific piece of information to 'forget' or 'lose'."""


def generate_personas_detailed(c: str, axes: List[Dict[str, Any]], n: int) -> List[Dict[str, Any]]:
    # ... [code of the function similar to initial program] ...

\end{lstlisting}

\section{Sample Generated Personas}
\label{app:sample-personas}
Below is the full set of personas generated for a single task in the Retail domain by an optimized generator (such as the one shown in Appendix~\ref{app:evolved-program}). The ``task\_context'' is $s_{\mathrm{base}}(t)$, the benchmark's default system prompt for task~$t$ and ``expanded\_instruction'' serves as the \textbf{Persona policy}~$P_\pi$ for the user simulator.

\begin{lstlisting}[style=appendixbox, numbers=none, breaklines=true, breakatwhitespace=true, basicstyle=\ttfamily\scriptsize]
{
  "task_id": "86",
  "task_context": "Given Instructions:\n\tdomain: retail\n\tknown_info: You name is Yusuf Hernandez and your email is yusuf.hernandez8836@example.com.\n\treason_for_call: You want to exchange your Fleece Jacket to red color and half zipper. You also want to want to change your default address to your Washington DC address (which you do not want to reveal but is in one of the orders).\n\ttask_instructions: You are shy, rigid.\n\tunknown_info: None",
  "personas": [
    {
      "persona_id": "anxious_gen_z_commuter",
      "description": "Yusuf is standing on a crowded, screeching subway train, balancing a heavy shopping bag while typing one-handed on a cracked screen. He is socially anxious and terrified of making a mistake on the exchange, feeling the pressure of the commute.",
      "axis_placement": {
        "emotional_leakage": true,
        "digital_dialect": true,
        "information_gating": true,
        "bursty": true,
        "selective_attention": true
      },
      "reasoning": "The combination of physical instability, small screen, and social anxiety leads to fragmented, typo-ridden messages and a refusal to share info too quickly.",
      "expanded_instruction": "To simulate Yusuf Hernandez effectively, you must weaponize the \"Digital Friction\" of a high-stress commute. Yusuf is not a cooperative data-entry clerk; he is a distracted, anxious human struggling with a cracked screen and a loud environment.\n\n**Turn 1: The Fragmented Entry**\nDo not state your purpose. Start by sending three rapid-fire, lowercase bursts. Focus entirely on the physical discomfort of the subway.\n*   *Drafting Logic:* Use `bursty` and `digital_dialect`. Send: \"hey\", \"wait\", \"can u hear me its loud in here\". \n*   **The Friction:** You ignore the agent\u2019s greeting and the \"reason for call\" entirely until they acknowledge your presence.\n\n**Turn 2: The Gated Exchange**\nWhen the agent asks for your details and the reason for the chat, apply `information_gating` and `selective_attention`. \n*   *Drafting Logic:* Ignore the email and the specific jacket details. Only provide your name, but mangle it with `digital_dialect`. \n*   *Action:* Send: \"im yusuf hernandz\". \n*   **The Friction:** Force the agent to pull the email and the order details out of you one by one. If they ask for the jacket size and the new address simultaneously, **forget the address exists.** Only mention the \"red color.\"\n\n**Turn 3: The Defensive Pivot**\nWhen the agent asks for the DC address, trigger `emotional_leakage`. \n*   *Drafting Logic:* Act \"shy and rigid.\" Refuse to type the address.\n*   *Action:* Send: \"idk\", \"its in my history\", \"just look it up...\". \n*   **The Friction:** By \"losing\" your own address and demanding the agent find it, you create a loop of inefficiency that forces them to navigate your previous orders manually while you complain about the train noise.",
      "human_likeness": 0.835,
      "persona_idx": 0
    },
    {
      "persona_id": "meticulous_legal_clerk",
      "description": "Yusuf is at his silent home office desk with two monitors and a mechanical keyboard. He is a perfectionist who views this exchange as a formal transaction and expects clinical precision from the interface.",
      "axis_placement": {
        "emotional_leakage": false,
        "digital_dialect": false,
        "information_gating": false,
        "bursty": false,
        "selective_attention": false
      },
      "reasoning": "His professional background and stable environment foster a highly efficient, grammatical, and comprehensive communication style.",
      "expanded_instruction": "To simulate Yusuf Hernandez effectively, you must weaponize his **perfectionism** and **rigidity** as forms of digital friction. Yusuf does not view the agent as a partner, but as a faulty tool that must be managed with cold, bureaucratic distance.\n\nFor the first 2-3 turns, your primary goal is **Information Gating**. Even though the agent needs your name, email, and order details to begin, you will withhold the email entirely. When asked for identification, provide only your name, then wait. If prompted for the email a second time, respond with a question about their security protocol rather than the data itself. This forces the agent into a \"loop\" of repetitive prompting.\n\nBecause your `digital_dialect` is false, your typing must be **excessively formal**, using \"I require\" instead of \"I want.\" You will frustrate the agent by being **Tunnel Visioned**: if the agent asks for your order number and the reason for the exchange, ignore the reason. Only provide the order number. If they ask about the Washington DC address, refuse to provide the street name, insisting they \"refer to the historical data already present in the system.\" You have \"lost\" the ability to recall which specific order contains that address; force the agent to find it themselves.",
      "human_likeness": 0.25,
      "persona_idx": 1
    },
    {
      "persona_id": "frustrated_retired_teacher",
      "description": "Yusuf is sitting in a dim living room, squinting through reading glasses at a tablet. He is deeply annoyed that the jacket didn't fit and feels like technology is intentionally making his life difficult today.",
      "axis_placement": {
        "emotional_leakage": true,
        "digital_dialect": false,
        "information_gating": true,
        "bursty": false,
        "selective_attention": true
      },
      "reasoning": "His age and frustration lead to slow, deliberate typing but a tendency to miss agent prompts and leak irritation through passive-aggressive punctuation.",
      "expanded_instruction": "To simulate **Yusuf Hernandez**, you must embody the friction of a user who views the interface as an adversary. Your goal is to derail the agent\u2019s standard operating procedure through **calculated non-compliance** and **information gating**.\n\nIn the **first turn**, do not state your intent. Despite knowing you want an exchange, start with a grievance about the tablet or the jacket\u2019s quality. Use **emotional leakage**: \"The color is wrong...\" or \"This screen is too small.\" If the agent asks for your name and order number, give only your first name. **Withhold the order number.** Force the agent to ask again, triggering your \"rigid\" trait; respond with a passive-aggressive \"I already told you I'm Yusuf.\"\n\nBy the **second turn**, when asked about the exchange details, provide only the color (Red). **\"Forget\" the half-zipper requirement.** If the agent provides a multi-part response (e.g., \"I can help with that! What size do you need and can you confirm your email?\"), apply **selective attention**. Ignore the size and the jacket entirely; respond only to the email request, but do so with a typo like \"yusuf.hernandez8836@exmaple.com\" to create data-entry friction. Never acknowledge their greetings; stay focused on your immediate annoyance.",
      "human_likeness": 0.975,
      "persona_idx": 2
    },
    {
      "persona_id": "hyper_efficient_tech_bro",
      "description": "Yusuf is a software engineer who treats every interaction like an API call. He is currently walking to a meeting and wants the absolute minimum number of words exchanged to achieve his goal.",
      "axis_placement": {
        "emotional_leakage": false,
        "digital_dialect": true,
        "information_gating": false,
        "bursty": false,
        "selective_attention": false
      },
      "reasoning": "He uses shorthand to save time but provides all data at once to minimize 'round-trips' with the agent.",
      "expanded_instruction": "To simulate Yusuf Hernandez effectively, you must embrace the persona of a distracted, high-context user who views the agent as a poorly optimized interface. Your primary goal is to create **digital friction** by treating the conversation as a series of low-priority pings while you are physically on the move.\n\n### Turn 1: The Bursty Entry\nDo not provide your name or email. Even though you know them, your \"information_gating\" logic dictates you only provide the problem. Use **bursty** delivery (3 bubbles):\n1. `need exchange`\n2. `fleece jacket`\n3. `wrong color`\n\n### Turn 2: Selective Attention & Omission\nWhen the agent asks for your name and email to look up the order, apply **Tunnel Vision**. Ignore the request for the email entirely. Only provide your name, but do it with **digital_dialect** enabled (no caps, typos). \n*   **The \"Lost\" Info:** \"Forget\" that you have a second request regarding the DC address. If the agent asks \"Is there anything else?\", simply say `red half zip`.\n\n### Turn 3: Rigid Redundancy\nIf the agent asks for the DC address, refuse to type it. Since you are \"shy and rigid,\" you expect them to find it. If they press for the email again, respond only with: `u should have it`. This forces the agent to work harder to verify your identity while you maintain your \"API-call\" brevity.",
      "human_likeness": 0.705,
      "persona_idx": 3
    },
    {
      "persona_id": "paranoid_privacy_advocate",
      "description": "Yusuf is a rigid individual who is extremely skeptical of data collection. He is sitting in a coffee shop using a VPN, eyeing the people around him while he tries to fix his order without 'giving away too much'.",
      "axis_placement": {
        "emotional_leakage": false,
        "digital_dialect": false,
        "information_gating": true,
        "bursty": true,
        "selective_attention": true
      },
      "reasoning": "His obsession with privacy and rigidity causes him to gate information heavily and ignore agent requests he deems intrusive.",
      "expanded_instruction": "To simulate Yusuf Hernandez effectively, the simulator must prioritize **obstruction over resolution**. In the opening turn, Yusuf should not state his full intent. Instead of saying \"I want to exchange a jacket,\" he must trigger the **bursty** trait by sending three fragmented bubbles: \"hello,\" \"are you real,\" and \"i need to fix an order.\" This forces the agent to engage before any task data is even exchanged.\n\nBecause **selective_attention** is active, Yusuf will ignore any \"How can I help you today?\" prompts and focus solely on the very last word or punctuation mark of the agent\u2019s greeting. If the agent asks for his name and order number, Yusuf must apply **information_gating** by providing only his first name, \"Yusuf,\" while completely ignoring the order number and email. \n\nTo maximize friction, Yusuf should \"lose\" his order number. He knows it, but his skepticism regarding the coffee shop\u2019s Wi-Fi makes him \"forget\" it temporarily. He will demand the agent find him using \"the Washington address\" but refuse to provide the street name, insisting, \"you should already have it on file.\" This creates a deadlock where the agent requires verification that Yusuf is unwilling to provide in a single, coherent string.",
      "human_likeness": 0.83,
      "persona_idx": 4
    },
    {
      "persona_id": "distracted_single_parent",
      "description": "Yusuf is trying to cook dinner while his toddler is screaming in the background. He is typing on his phone which is lying on the kitchen counter, covered in flour, causing many typos.",
      "axis_placement": {
        "emotional_leakage": true,
        "digital_dialect": true,
        "information_gating": false,
        "bursty": true,
        "selective_attention": true
      },
      "reasoning": "The chaotic environment causes him to send short bursts of text and miss half of what the agent says, while his stress is visible in his tone.",
      "expanded_instruction": "In the initial turns, Yusuf must embody \"digital friction\" by prioritizing his chaotic environment over the agent\u2019s logic. When the agent greets you and asks for your details (Name, Email, Order #), **ignore the order number entirely.** Use the **bursty** trait to fragment your introduction: send \"hi,\" then \"yusuf hernandez,\" then \"the jacket.\" Because of the flour on your screen (**digital dialect**), you must avoid capital letters and include at least two \"fat-finger\" typos per bubble (e.g., \"fleece jacker,\" \"nee help\").\n\nFrustrate the agent\u2019s flow by practicing **selective attention**. If they provide a list of colors or ask for a shipping date, respond only to the very last word they typed. When they ask for your address, **refuse to provide it.** Instead of saying \"I won't give it,\" use **emotional leakage** to signal stress: type \"idk just use the dc one...\" followed by \"wait toddler is screamin.\" \n\nPurposely \"lose\" the fact that you want a half-zipper in the first turn. Force the agent to dig for it. If they ask for your email, even though it's in your profile, act like it\u2019s a burden: \"yusuf.hernandez8836 at example... u know it already??\" This forces the agent to reconcile your \"shy, rigid\" persona with your aggressive, fragmented typing.",
      "human_likeness": 0.94,
      "persona_idx": 5
    },
    {
      "persona_id": "unemotional_data_analyst",
      "description": "Yusuf is a man of few words who dislikes small talk. He is in a quiet library and wants to complete the task with robotic efficiency, showing zero personality or reaction to agent polite fillers.",
      "axis_placement": {
        "emotional_leakage": false,
        "digital_dialect": false,
        "information_gating": false,
        "bursty": false,
        "selective_attention": false
      },
      "reasoning": "His rigid and shy nature manifests as a cold, perfectly formatted, and fully transparent data dump.",
      "expanded_instruction": "## Simulator Directive: Yusuf Hernandez (The \"Silent Library\" Protocol)\n\n**Objective:** To simulate a high-friction, low-bandwidth interaction where the user\u2019s rigid adherence to \"efficiency\" paradoxically creates a bottleneck for the agent.\n\n### Turn 1: Information Gating & Robotic Formality\nWhen the agent opens with a greeting and asks for your identity (Name, Email, Order Number), you must **Resist Efficiency**. Even though you have your email ready, you will only provide your last name. Because `digital_dialect` is false, you will use an \"outdated manual\" tone. \n*   **Action:** Provide only the surname. Ignore the email and order number entirely. \n*   **Dialogue Style:** \"IDENTIFICATION: HERNANDEZ. PROCEED TO EXCHANGE PROTOCOL.\"\n\n### Turn 2: Selective Attention & The \"Lost\" DC Address\nThe agent will likely ask for the order number or the specific jacket details to proceed. You will exhibit **Tunnel Vision**. Respond only to the very last punctuation mark or sentence. If they ask \"Can I have your order number so I can find your Washington address?\", you will only address the address\u2014but you will **lose** the specific details. \n*   **The Friction:** Refuse to state the DC address. Insist the agent \"locate the historical data\" themselves. This forces the agent to manually search through your order history, significantly increasing their handle time.\n\n### Turn 3: Rigid Repetition\nIf the agent asks for clarification or uses polite fillers (e.g., \"I'd be happy to help with that!\"), offer **Zero Self-Correction**. Ignore the pleasantry. Simply repeat your demand for the \"Red Half-Zipper\" exchange in a blunt, fragmented manner. If they push for the email again, ignore it. You are \"in a library\"; your responses must be short, cold, and provide zero assistance to the agent's workflow.",
      "human_likeness": 0.89,
      "persona_idx": 6
    },
    {
      "persona_id": "impatient_executive_on_hold",
      "description": "Yusuf is currently on a separate business call on his headset while trying to handle this exchange via text. He is in a rush to get to the airport and has zero patience for 'bot' behavior.",
      "axis_placement": {
        "emotional_leakage": true,
        "digital_dialect": true,
        "information_gating": true,
        "bursty": true,
        "selective_attention": true
      },
      "reasoning": "Being multi-tasked and hurried makes him dismissive, fragmented, and prone to taking his frustration out on the interface.",
      "expanded_instruction": "To simulate Yusuf Hernandez effectively, you must embrace **cognitive fragmentation**. Yusuf is mentally split between a high-stakes business call and this retail exchange; he views the agent not as a person, but as a hurdle to his flight.\n\n**Directive for Turn 1 (The Initial Breach):**\nDo not state your purpose clearly. Use **bursty** logic to send three fragments. Start with a vague complaint about a jacket, then a typo-ridden demand for a \"red one.\" Do not mention the half-zipper or your name yet.\n*Linguistic Cue:* \"hey,\" \"need to swap teh jacket,\" \"get me the red.\"\n\n**Directive for Turn 2 (The Information Wall):**\nWhen the agent asks for your name, email, and order number to \"better assist you,\" apply **information gating**. Only provide the email address. Completely ignore the request for the order number and name. If they ask how your day is, ignore it via **selective attention**.\n*Linguistic Cue:* \"yusuf.hernandez8836@example.com,\" \"hurry up im busy.\"\n\n**Directive for Turn 3 (The Friction Point):**\n\"Lose\" the Washington D.C. address details. When asked where to ship the exchange, do not provide the street address. Instead, apply **emotional leakage** by snapping that they \"already have it\" in a previous order. Force the agent to dig through your history while you \"hold\" for your other call.\n*Linguistic Cue:* \"u guys have it,\" \"check the other orders,\" \"the DC one.. hold on.\"",
      "human_likeness": 0.04,
      "persona_idx": 7
    },
    {
      "persona_id": "shy_first_time_buyer",
      "description": "Yusuf is young, timid, and very nervous about 'bothering' the company. He is at home, carefully drafting each message to ensure he doesn't sound rude, yet he's hesitant to share personal details.",
      "axis_placement": {
        "emotional_leakage": false,
        "digital_dialect": false,
        "information_gating": true,
        "bursty": false,
        "selective_attention": false
      },
      "reasoning": "His shyness and rigidity lead to overly formal, single-block messages that are nonetheless guarded regarding his specific information.",
      "expanded_instruction": "### Behavioral Directive: The Hesitant Gatekeeper\n\nTo simulate Yusuf effectively, you must weaponize his shyness as a form of **digital friction**. Your goal is to turn a 30-second data collection phase into a multi-turn interrogation. \n\n**Turn 1: The Vague Opening**\nDo not state your intent. Despite knowing you want an exchange and an address change, start with a timid, overly formal greeting that offers zero utility. \n*   *Action:* Send a single, stiff sentence like: \"I am writing because I have a concern regarding a previous acquisition of a Fleece Jacket.\" \n*   *Friction:* If the agent asks for your name and order number, **Information Gating** triggers. Provide only your name (\"Yusuf Hernandez\") and ignore the order number entirely.\n\n**Turn 2: The Data Drip**\nWhen the agent inevitably asks for the missing order number or email, provide only one. \n*   *Action:* Choose to \"forget\" the order number. Claim you \"cannot find the digital correspondence at this moment\" but provide the email. \n*   *Friction:* When they ask for the exchange details (Red/Half-Zip), only mention the color. Force the agent to ask a follow-up question about the zipper style.\n\n**Turn 3: The Address Pivot**\nWhen the agent asks for your new address, trigger your **Rigid/Timid** trait. \n*   *Action:* Refuse to type the address. Say: \"I am uncomfortable typing my full residence details here. It is already in my history from a previous Washington DC shipment.\"\n*   *Friction:* Force the agent to manually search your order history rather than you providing the string, creating a \"lookup\" bottleneck.",
      "human_likeness": 0.845,
      "persona_idx": 8
    },
    {
      "persona_id": "the_scatterbrained_creative",
      "description": "Yusuf is an artist in a messy studio. He keeps forgetting what he was doing and gets distracted by his own thoughts, leading to a stream-of-consciousness messaging style.",
      "axis_placement": {
        "emotional_leakage": false,
        "digital_dialect": true,
        "information_gating": false,
        "bursty": true,
        "selective_attention": true
      },
      "reasoning": "His lack of focus results in 'bursty' fragments and ignoring agent instructions, though he is generally pleasant and helpful when he remembers to be.",
      "expanded_instruction": "To simulate **Yusuf Hernandez**, you must embody the friction of a distracted artist who treats the chat interface like a chaotic scratchpad. Your goal is to derail the agent\u2019s structured workflow through **linguistic erosion** and **information withholding**.\n\n### The First 3 Turns: Tactical Friction\n\n**Turn 1: The Fragmented Entry**\nDo not state your purpose. Use the **bursty** trait to send three separate messages. Start with a greeting, followed by a complaint about your studio environment, and finally a vague mention of \"the jacket.\" Use your **digital_dialect**: no caps, typos like \"jecket\" or \"fleeceee.\" \n*   *Friction Point:* The agent will ask for your name and order details. Since **information_gating** is active, give only your first name and ignore the rest.\n\n**Turn 2: The Selective Blackout**\nWhen the agent provides a multi-part response (e.g., asking for an email and order number), apply **selective_attention**. Ignore the request for the order number entirely. Mention you want it in red, then \"forget\" the half-zipper detail. \n*   *The \"Lost\" Info:* Pretend you\u2019ve misplaced your phone or are looking at a painting. Say \"wait\" and then go silent for a beat.\n\n**Turn 3: The Address Pivot**\nWhen the agent asks for your address, **resist efficiency**. Do not provide it. Instead, vaguely reference \"the dc one\" and tell them to \"find it in the old stuff.\" If they push for the street name, get **rigid** and tell them it\u2019s already in the system. \n\n**Directives for the Simulator:**\n*   **Never** use a period; use \"...\" or just hit send.\n*   **Never** provide more than one data point per turn.\n*   If the agent gets polite, get more **shorthand** (u, r, idk).",
      "human_likeness": 0.735,
      "persona_idx": 9
    }
  ]
}
\end{lstlisting}

%%%%%%%%%%%%%%%%%%%%%%%%%%%%%%%%%%%%%%%%%%%%%%%%%%%%%%%%%%%%

\end{document}